\documentclass[journal]{IEEEtran}  
\usepackage{graphicx}
\usepackage{amsmath,amssymb}
 \usepackage{multirow} 
 \usepackage{subcaption}
\usepackage{booktabs}
\usepackage{makecell}

\usepackage{booktabs}
\usepackage{makecell}
\usepackage{siunitx}
\IEEEoverridecommandlockouts                              

\title{\LARGE \bf
Generating Roadside LiDAR Datasets from Vehicle-Side Datasets via Novel View Synthesis
}

\author{
Yuhan~Xia$^{\dagger}$,
Runxin~Zhao$^{\dagger}$,
Hanyang~Zhuang,~\IEEEmembership{Member,~IEEE},\\
Chunxiang~Wang,
and Ming~Yang,~\IEEEmembership{Member,~IEEE}%
\thanks{This work was supported by National Natural Science Foundation of China
(62573295/U22A20100/62373250).
Hanyang Zhuang and Ming Yang are the corresponding authors.}%
\thanks{$^{\dagger}$Yuhan Xia and Runxin Zhao contributed equally to this work.}%
\thanks{Yuhan Xia, Runxin Zhao, Chunxiang Wang and Ming Yang are with the School of Automation and Intelligent Sensing,
Shanghai Jiao Tong University, Shanghai 200240, China;
Key Laboratory of System Control and Information Processing,
Ministry of Education of China, Shanghai 200240, China
(e-mail: mingyang@sjtu.edu.cn).}%
\thanks{Hanyang Zhuang is with the Global College,
Shanghai Jiao Tong University, Shanghai 200240, China
(e-mail: zhuanghany11@sjtu.edu.cn).}%
}

\begin{document}

\maketitle
\thispagestyle{empty}
\pagestyle{empty}

\begin{abstract}


Intelligent Transportation Systems (ITS) require reliable environmental perception to support safe and efficient transportation.
With the rapid development of Vehicle-to-everything (V2X), roadside perception has become an effective means to extend sensing coverage and improve traffic safety. However, the scarcity of large-scale annotated roadside LiDAR datasets poses a major challenge for training high-performance roadside perception models. In this paper, we introduce Vehicle-to-Roadside LiDAR Synthesis (VRS), a data synthesis framework that generates labeled roadside LiDAR datasets from vehicle-side datasets via LiDAR novel view synthesis. To mitigate the vehicle-to-roadside domain gap, VRS employs vehicle point cloud completion to compensate for missing geometry in vehicle-side
observations, and introduces an occupancy-based visibility constraint to handle large viewpoint changes during cross-view rendering. The proposed framework enables flexible multi-view rendering for scalable roadside data generation. Extensive experiments on roadside 3D object detection
demonstrate that the synthesized data effectively complements
real roadside data, mitigates the limitations of limited
real-world roadside data, and improves generalization to unseen
roadside viewpoints.


\end{abstract}

\section{INTRODUCTION}


In Intelligent Transportation Systems (ITS), Vehicle-to-everything (V2X) supports information sharing among connected entities. As a key component of V2X, roadside perception extends the perception range of individual vehicles and enriches decision-making information, thereby significantly enhancing traffic safety. However, the scarcity of public roadside datasets limits the  development of high-performance roadside perception models. Meanwhile, the autonomous driving community has accumulated a wealth of finely annotated vehicle-side datasets, such as KITTI \cite{geiger2013vision}, Waymo \cite{sun2020scalability}, and nuScenes \cite{caesar2020nuscenes}. Inspired by this, we explore LiDAR novel view synthesis to generate roadside datasets from existing real-world, well-annotated vehicle-side data. This approach not only preserves the high-fidelity of both scenes and vehicles but also greatly reduces the cost of data collection and annotation.

 
 Previously, LiDAR novel view synthesis has been used to generate new viewpoint data near the original vehicle perspective, thereby enhancing data diversity.  Traditional methods \cite{manivasagam2020lidarsim,li2022pcgen} relied on explicit geometric reconstructions of scenes, but these approaches were often inaccurate, noisy, and limited to static environments. 
More recent methods adopt implicit neural representations, significantly improving modeling fidelity for LiDAR-based novel view synthesis \cite{tao2024lidar,huang2023neural,zheng2024lidar4d,wu2024dynamic}.  
However, these methods are still limited to synthesizing viewpoints near the original perspective.


 To address the challenges of large viewpoint changes from vehicle-side to roadside, we introduce Vehicle-to-Roadside LiDAR Synthesis (VRS), a novel framework for synthesizing high-fidelity and realistic roadside LiDAR datasets from existing vehicle-side datasets. VRS first decomposes the vehicle-side point cloud into static background and dynamic vehicles. For dynamic vehicles, quality filtering and point cloud completion are applied to recover missing geometric structures and aggregate complete multi-view observations. For the background, unannotated dynamic objects are removed and global pose alignment is performed in the world coordinate system to ensure geometric consistency. Dedicated neural fields are then constructed for the static background and individual vehicles, respectively. Finally, under custom roadside LiDAR poses, these neural fields are jointly rendered with occupancy-based visibility constraints, producing multi-view synthetic roadside point clouds with accurate 3D annotations that directly support downstream roadside perception tasks.

Our VRS framework achieves high-fidelity vehicle-to-roadside LiDAR novel view synthesis under large viewpoint changes. Extensive experiments on the V2X-Seq dataset \cite{yu2023v2x} demonstrate the
effectiveness of VRS. Results on roadside 3D object
detection show that synthetic roadside LiDAR data consistently
outperforms direct vehicle-side training and effectively
complements real roadside data. The performance gains are
most pronounced under limited real-world roadside data and
translate to improved generalization to unseen roadside
viewpoints. 

In summary, our main contributions are:

 \begin{itemize}


\item We present Vehicle-to-Roadside LiDAR Synthesis (VRS), an innovative data synthesis framework based on composite neural fields that generates realistic and labeled roadside LiDAR point clouds from vehicle-side datasets. By enabling scalable roadside data generation, VRS effectively facilitates the training of roadside perception models.


 \item We mitigate the vehicle-to-roadside domain gap by introducing a vehicle point cloud completion pipeline that compensates for missing geometry in vehicle-side observations.


 \item  We introduce an occupancy-based visibility constraint for large cross-view viewpoint changes, which suppresses unreliable geometry predictions in unobserved regions.


\end{itemize}




The remainder of this paper is organized as follows. Section II reviews related work.  Section III presents the proposed VRS framework. Section IV reports experimental results on the V2X-Seq dataset \cite{yu2023v2x}, and Section V concludes the paper.

\section{Related Work}
\begin{figure*}[ht]
  \centering
  \includegraphics[width=1\textwidth]{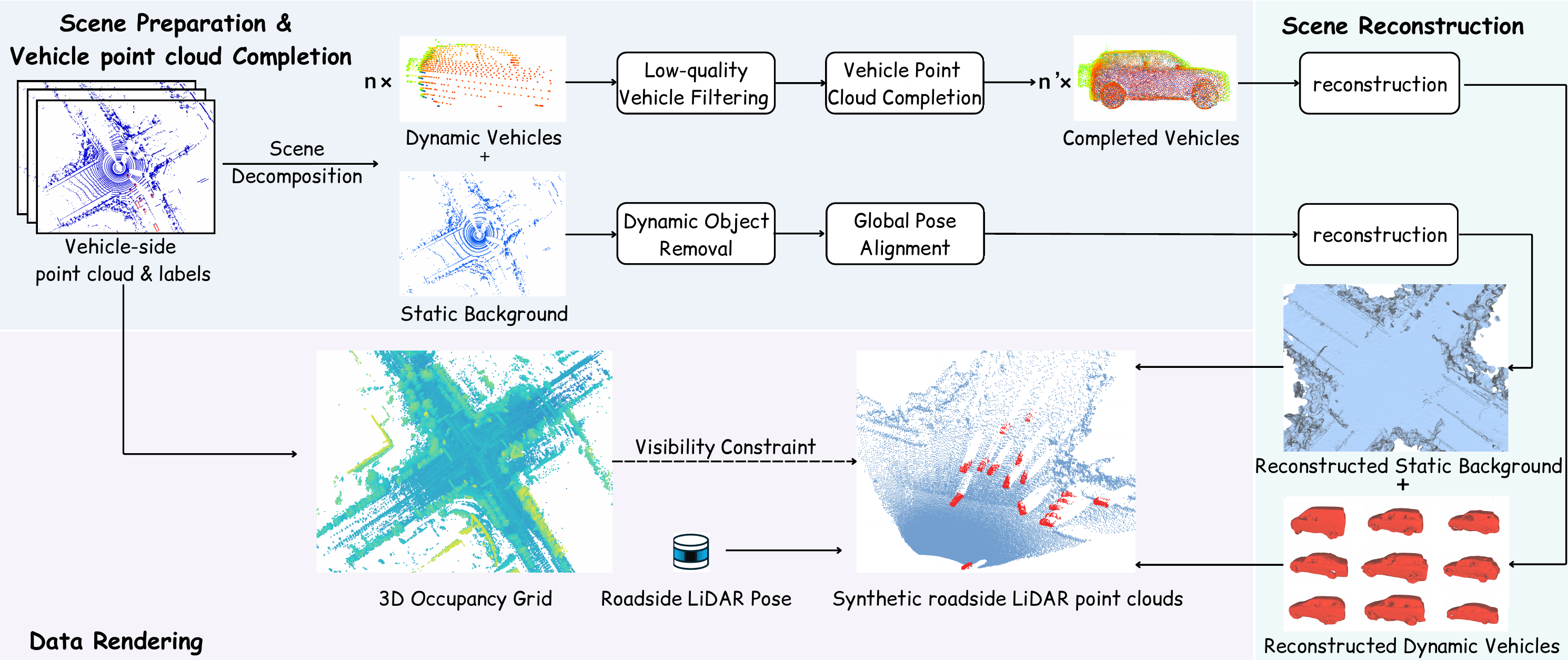}
  \caption{Overview of VRS. Our method takes annotated vehicle-side point cloud data as input. VRS first decomposes the scene into a static background and dynamic vehicles. Vehicle instances with insufficient observations are filtered out, and the remaining vehicles are completed to recover full geometry. The static background is processed by dynamic object removal and global pose alignment. Both components are reconstructed using neural fields and rendered from target roadside LiDAR viewpoints under 3D occupancy constraints, producing high-fidelity synthetic roadside LiDAR point clouds.}

  \label{fig:pdf-example}
 \end{figure*}
 \subsection{Roadside Dataset}

 Obtaining roadside datasets is fundamental for enhancing the performance of roadside perception systems.  Currently, there are two main approaches to collect roadside datasets.  The first method is to generate  virtual datasets \cite{xu2022v2x,li2022v2x,mao2022dolphins,wang2024deepaccident}  using simulation tools.  For example, V2XSet \cite{xu2022v2x}  and V2X-Sim \cite{li2022v2x} leverage synthesized data from the CARLA simulator \cite{dosovitskiy2017carla}  to explore V2X perception.  While this approach allows for the rapid generation of large-scale datasets, it is limited by the domain gap between the simulated data and real-world scenarios, which restricts the model's performance when applied to real-world environments.  The second method involves the collection of real-world roadside data. DAIR-V2X \cite{yu2022dair} introduces the first real-world dataset for cooperative detection.  RCooper \cite{hao2024rcooper} expands from a single roadside sensor to multiple roadside sensors, further advancing roadside cooperative perception.  Although this method provides data with high authenticity and temporal relevance, it incurs high data collection and annotation costs, making it challenging to acquire sufficient data.  To address these challenges, we propose a novel method that generates roadside datasets from vehicle-side data using LiDAR novel view synthesis.  This approach not only enables the generation of high-fidelity, real-world-like data but also significantly reduces costs, thereby offering a more accessible and scalable solution for training roadside perception models.

\subsection{Novel View Synthesis}

To reduce data collection and annotation costs while preserving realism,
recent studies explore novel view synthesis by reconstructing scenes
from recorded sensor data and re-rendering observations from new viewpoints.
Existing LiDAR novel view synthesis methods can be broadly categorized into
explicit and implicit approaches.

Early works, such as LiDARsim~\cite{manivasagam2020lidarsim} and PCGen~\cite{li2022pcgen}, rely on explicit geometric representations to reconstruct the scene.  To bridge the simulation-to-reality gap, these methods further train networks to replicate the physical characteristics of LiDAR light projection.  However, these explicit approaches often suffer from lower accuracy, increased noise, and are primarily suited for static scene reconstruction.

The introduction of LiDAR-NeRF~\cite{tao2024lidar} and NFL~\cite{huang2023neural} marked a significant step forward, employing implicit neural field modeling for LiDAR novel view synthesis.  Despite their contributions, these methods remain constrained to static scene reconstructions.  In response to the challenges of dynamic scene reconstruction, Lidar4D~\cite{zheng2024lidar4d} introduced 4D hybrid neural representations and motion priors derived from point clouds, achieving strong performance in large-scale dynamic scene reconstruction.
 Nevertheless, it does not explicitly decouple foreground and background,
which limits its flexibility for scene editing and composition.
  DyNFL~\cite{wu2024dynamic} proposed compositional neural fields that decouple foreground and background modeling, facilitating flexible scene editing and composition. It further integrates reconstructed neural assets from diverse scenes through ray drop test, effectively accounting for occlusions and transparent surfaces, leading to significant improvements in dynamic scene LiDAR simulation.
  
 However, the aforementioned methods are only suitable for novel view synthesis near the original perspective and cannot adapt to large angular changes from the vehicle-side to the roadside, especially leading to poor quality of the generated vehicle point clouds. Our work is the first to apply LiDAR novel view synthesis for the generation of roadside datasets.

 \subsection{Point Cloud Completion}

 Existing point cloud completion methods can generally be divided into two categories: full-shape reconstruction and missing-part prediction.

Full-shape reconstruction methods generate a uniformly distributed complete point cloud from scratch by extracting global features. The pioneering work PCN \cite{yuan2018pcn} adopts an encoder-decoder framework to produce a coarse output and then refine local details. Building on this idea, later approaches \cite{xie2020grnet,wang2021voxel,huang2021rfnet,liu2020morphing,pan2021variational,wang2020cascaded,tchapmi2019topnet} improve completion performance through enhanced feature extraction strategies. More recently, Transformer-based models \cite{li2023proxyformer,fu2023vapcnet,zhou2022seedformer} incorporate attention mechanisms \cite{vaswani2017attention} to better capture fine-grained geometric structures. However, these
methods often discard the rich information from the original
point cloud input, resulting in generated point clouds lacking
detail.

In contrast, missing-part prediction methods estimate only the absent regions and concatenate them with the observed input. PF-Net \cite{huang2020pf} employs a multi-scale feature-point generation scheme for progressive completion, while PoinTr \cite{yu2021pointr} leverages Transformers to predict proxy points for missing segments. Although such methods largely preserve the original geometry, they frequently suffer from inconsistencies at the boundary between existing and reconstructed parts. 

To achieve global consistency optimization while retaining the original structural details, SymmCompletion \cite{yan2025symmcompletion} first reconstructs the missing parts and then applies a global optimization network to refine the initial point cloud, achieving state-of-the-art performance.

 \section{Method}

VRS aims to generate high-fidelity roadside LiDAR datasets from 
well-annotated vehicle-side datasets using LiDAR novel view 
synthesis. The overall pipeline is 
illustrated in Fig.~1. 

We organize the remainder of this section as follows.
Section III-A introduces vehicle point cloud completion and
quality filtering. Section III-B describes neural field
reconstruction for the static background and dynamic vehicles.
Section III-C presents the roadside data rendering process
with occupancy-based visibility constraint.

 \subsection{Vehicle Point Cloud Completion} 

\begin{figure*}[ht]
  \centering
  \includegraphics[width=1\textwidth]{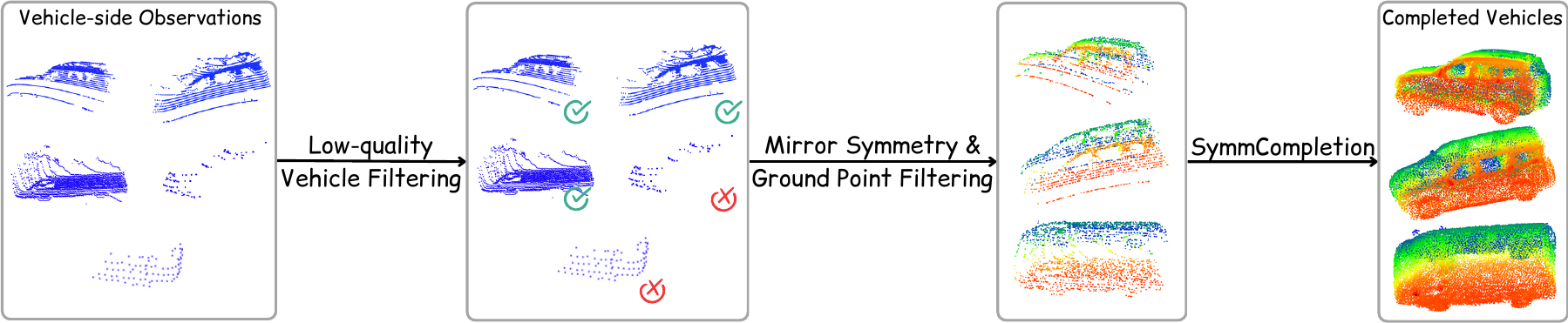}
  \caption{The pipeline of vehicle point cloud
completion.}
  \label{fig:pdf_com}
 \end{figure*}

 
 Since there is a significant difference between the vehicle-side and roadside perspectives, training a neural field solely on vehicle-side observations would result in missing details in parts that are not visible in the vehicle-side observations, such as the roof or rear of the vehicle. To enable the vehicle neural field to learn a complete representation of each vehicle, we apply point cloud completion to the extracted vehicle point clouds and use the completed results as inputs for subsequent reconstruction. 
Figure~\ref{fig:pdf_com} illustrates the overall pipeline of vehicle point cloud completion.

When the number of observed points is extremely limited, the completion network may generate unrealistic shapes that deviate from the true vehicle geometry. To prevent such failure cases, vehicles whose input point count falls below a predefined threshold are labeled as "unreconstructable" and excluded from the completion and reconstruction pipeline.

For each vehicle instance that passes the low-quality filtering step, 
we select the most informative observation as the completion input.
Specifically, we choose the single frame with the largest number of observed points, 
which provides the most reliable geometric evidence while avoiding geometric inconsistencies 
that may arise from multi-frame aggregation and adversely affect subsequent neural field reconstruction.
The selected point cloud is further preprocessed to improve completion quality.
Considering that vehicles typically exhibit approximately symmetric geometries, 
we first apply mirror symmetry augmentation to enrich the geometric cues available for completion.
Since vehicle point clouds extracted from bounding boxes may contain a small number of ground points, 
we then apply ground point filtering to prevent the completion network from mistakenly reconstructing 
ground structures as part of the vehicle.

Finally, we adopt a learning-based approach for vehicle point cloud completion.
We employ SymmCompletion~\cite{yan2025symmcompletion}, which achieves state-of-the-art performance in point cloud completion 
and is particularly effective when vehicle point clouds are severely incomplete.


 \subsection{Scene Reconstruction}

 We establish a static neural field \( F_{\text{static}} \) for the background and a set of dynamic neural fields for the $n'$ selected vehicles.  Rays intersecting any vehicle bounding box are used to train the corresponding vehicle field, whereas rays not intersecting any bounding box are assigned to the background field.

 \textbf{Static Background Reconstruction:}  The static background is encoded as a neural field \( F_{\text{static}}: (x, d) \mapsto (s, p_d) \), which predicts the signed distance \( s \) and ray drop probability \( p_d \in [0, 1] \) for a given ray with origin \( x \) and direction \( d \).  For \( F_{\text{static}} \), we optimize the neural field representation by minimizing the following loss:
 \begin{equation}
 L = w_\zeta L_\zeta + w_s L_s + w_{\text{eik}} L_{\text{eik}} + w_{\text{drop}} L_{\text{drop}}
 \end{equation}
 where \( L_\zeta \) is the L1 loss for distance estimation.  \( L_s \) is the surface point SDF regularization loss, which penalizes non-zero SDF values at surface points.  \( L_{\text{eik}} \) is the Eikonal constraint loss, which regularizes the SDF level set and ensures that the gradient norm of the SDF is close to 1.  \( L_{\text{drop}} \) is the ray drop loss, combining binary cross-entropy loss and Lovasz loss to supervise ray drop estimation, simulating the situation where rays are dropped due to insufficient return power.


Since the spatial coverage of a single vehicle-side sequence is typically insufficient for reconstructing the full roadside field of view, we jointly train a unified background neural field using all fragments from the same scene. This allows the model to learn a background representation that spans the entire environment. However, in practical multi-fragment data collection scenarios, vehicle-side LiDAR poses are often not globally optimized. When directly aligning different fragments into the world coordinate system, slight misalignments and ghosting appear, which become more pronounced when multiple fragments are combined. Such pose inconsistencies introduce conflicting supervision during training,
forcing the neural field to average over misaligned observations and resulting
in a blurred and ambiguous background representation. This representation
can propagate to downstream 3D object detection, weakening the negative evidence
needed to distinguish background structures from vehicles and leading to more
false positive detections.
To ensure globally consistent observations across fragments, we follow HBA \cite{liu2023large}, a globally consistent and efficient large-scale LiDAR mapping method, to optimize the vehicle-side poses.
Moreover, in many vehicle-side perception datasets, dynamic object annotations are typically limited to the ego vehicle’s forward field of view, leaving objects in side and rear regions unlabeled. These unannotated dynamic objects are mistakenly treated as background during training, introducing noise into the background field reconstruction. To obtain a clean and reliable background, we employ a 3D detection model trained on vehicle-side point clouds to generate pseudo-labels for these regions and filter out the corresponding vehicle points, ensuring that the background remains static and consistent.

 \textbf{Dynamic Vehicle Reconstruction:}  For each dynamic vehicle, we establish a canonical coordinate system based on the first frame.  Assuming that each dynamic vehicle undergoes only rigid motion, we reconstruct them statically in the canonical space and render their motion over time by inverting the neural field alignment.

 Specifically, consider the point cloud sequence of a dynamic vehicle $v$ $\{X_t^v\}_{t=1}^{T}$, and its corresponding sequence of bounding boxes $\{B_t^v\}_{t=1}^{T}$, where $B_t^v \in \mathbb{R}^{3\times 8}$.  We treat $B_1^v$ as the \textit{canonical box}.  For the remaining $T-1$ boxes, we estimate transforms $\{T_t\}_{t=2}^{T}$ with $T_t\in SE(3)$ such that
 \begin{equation}
   B_1^v = T_t\, B_t^v, \quad \forall\, t\in\{2,\dots,T\}.
 \end{equation}

 The neural field $F_v$ for vehicle $v$ is reconstructed in the canonical space, following the same neural field representation and loss function as the static neural field.  During rendering, the corresponding transformation $T_t$ is applied for coordinate conversion, formulated as
 \begin{equation}
     F_t^v : (T_t x, T_t d) \rightarrow (s, p_d).
 \end{equation}


 It is worth noting that the definition of ray drop in the dynamic vehicle neural field is broader than that in the static background field. While the background field treats only rays with insufficient return power as drops,
the dynamic field also includes rays that traverse the vehicle bounding box without
intersecting the actual vehicle surface. This additional drop category is essential for maintaining correct occlusion between the background and the vehicle fields. Without treating these rays as drops, the vehicle field would misinterpret them as surface hits merely because they pass through the bounding box, resulting in false geometry and incorrect foreground–background separation.
 


 The completed vehicle point clouds already provide sufficient multi-view information for novel view synthesis. Moreover, the training of an SDF field is highly sensitive to noise, and aligning multi-frame point clouds for training tends to introduce extra noise that degrades performance. Therefore, we use only the completed point clouds as input instead of fusing the entire sequence.

\begin{figure}[htbp]
   \centering
   \includegraphics[width=0.45\textwidth]{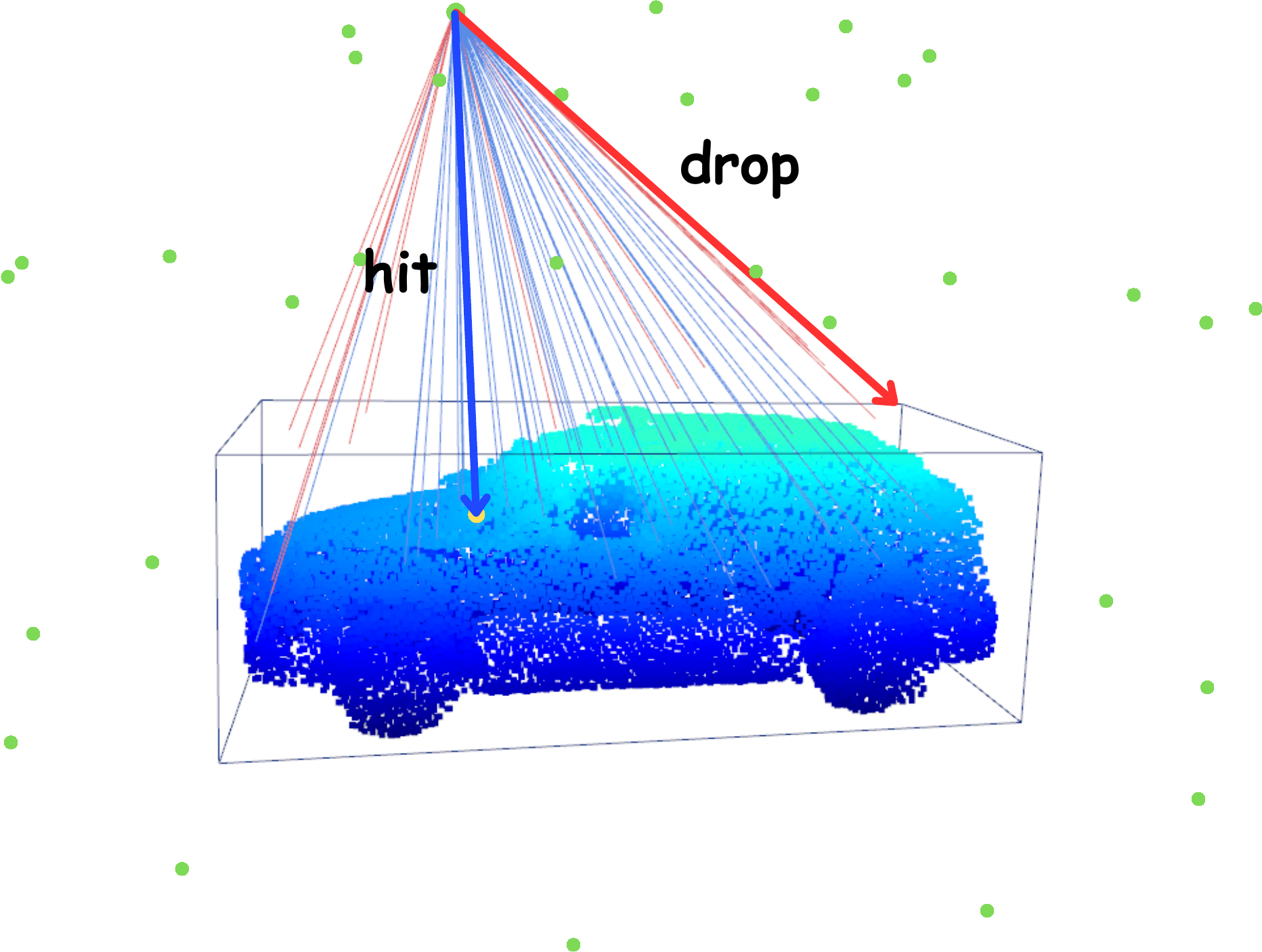}
   \caption{Illustration of the multi-view ray sampling strategy. Ray origins sampled around the vehicle are shown in green. Rays intersecting the vehicle are labeled as \emph{hit} (blue), while rays missing the object are labeled as \emph{drop} (red).}
   \label{fig:pic1}
\end{figure}

 
 To convert the completed point clouds into a ray-based representation
suitable for neural field reconstruction (\(x, d, p_d\)), we design a multi-view ray sampling strategy, as illustrated in Fig.~\ref{fig:pic1}.
For sufficient multi-view coverage, we arrange multiple layers of ring-shaped ray origins around the vehicle at different heights and radii, resulting in a uniformly distributed set of viewpoints. For each origin, we generate ray directions that pass through the vehicle region and classify them as hit or drop based on their minimum distance to the point cloud. A ray is considered a hit only when this distance is below 5 cm; otherwise, it is labeled as drop. Drop rays are explicitly used to supervise the ray drop probability, constraining
the model to learn accurate vehicle geometry and preventing hallucinated
structures in empty regions inside the bounding box.



 \begin{figure*}[htbp]
   \centering
   \includegraphics[width=1\textwidth]{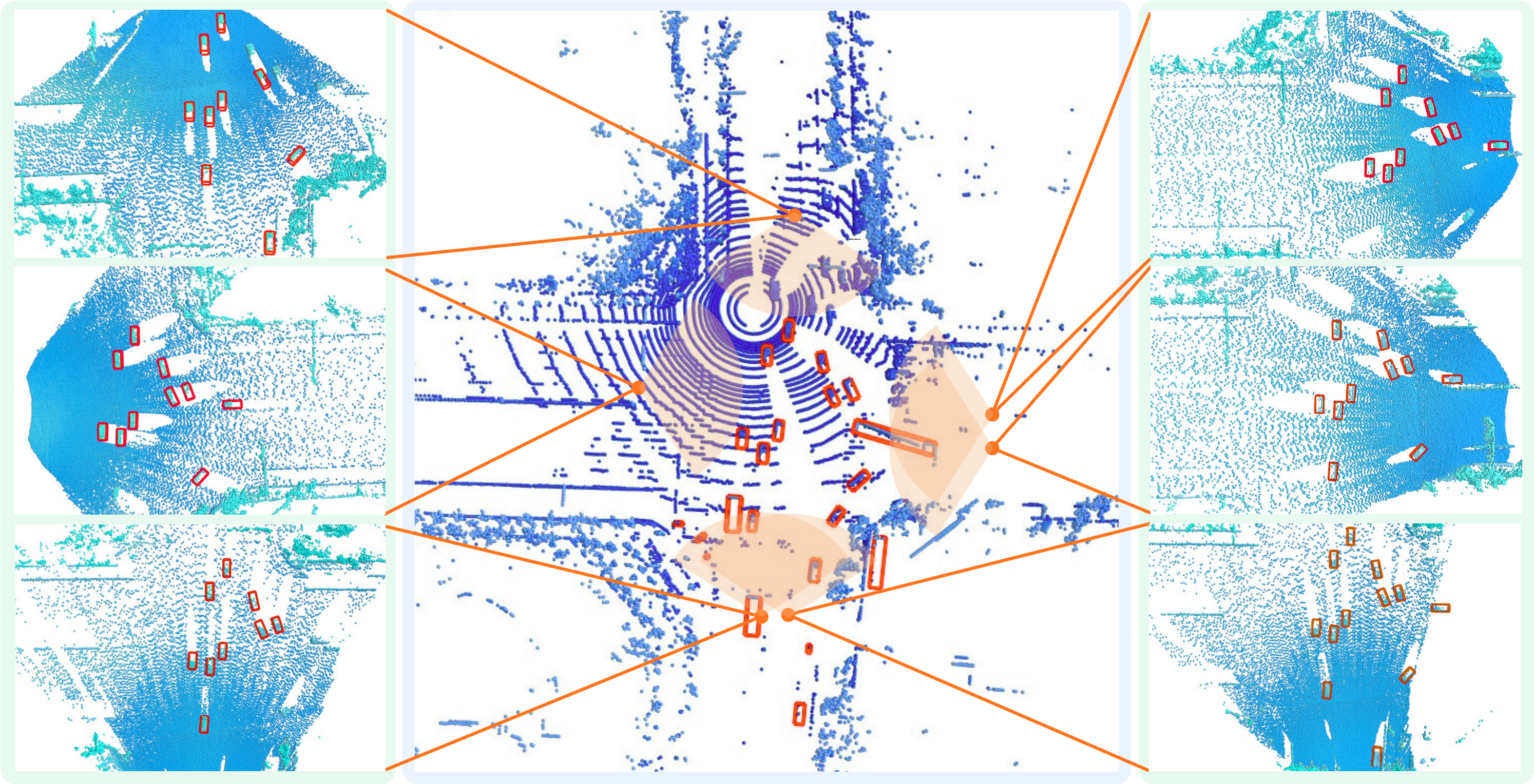}
   \caption{VRS generates point cloud data for 6 corresponding virtual roadside LiDAR poses based on vehicle-side LiDAR observation data.  The middle part of the figure shows the vehicle-side point cloud data, while the left and right sides display the generated roadside LiDAR point cloud data. The synthesized
roadside point clouds preserve clean geometric structures.}
   \label{fig:pdf-example3}
   
 \end{figure*}

 \subsection{Data Rendering}

When rendering from a roadside viewpoint, there exists a significant distribution mismatch between the vehicle-side training data and the target rendering views. Roadside LiDAR sensors are typically mounted at higher positions and offer wider fields of view, enabling them to capture distant road surfaces and large sky regions that are never observed by the vehicle-side sensor during data acquisition. As a result, rays cast from roadside viewpoints frequently traverse spatial regions without any training supervision, forcing the neural field to extrapolate and often leading to artifacts such as distorted distant road geometry or spurious sky structures.

\begin{figure}[h]
   \centering
   \includegraphics[width=0.47\textwidth]{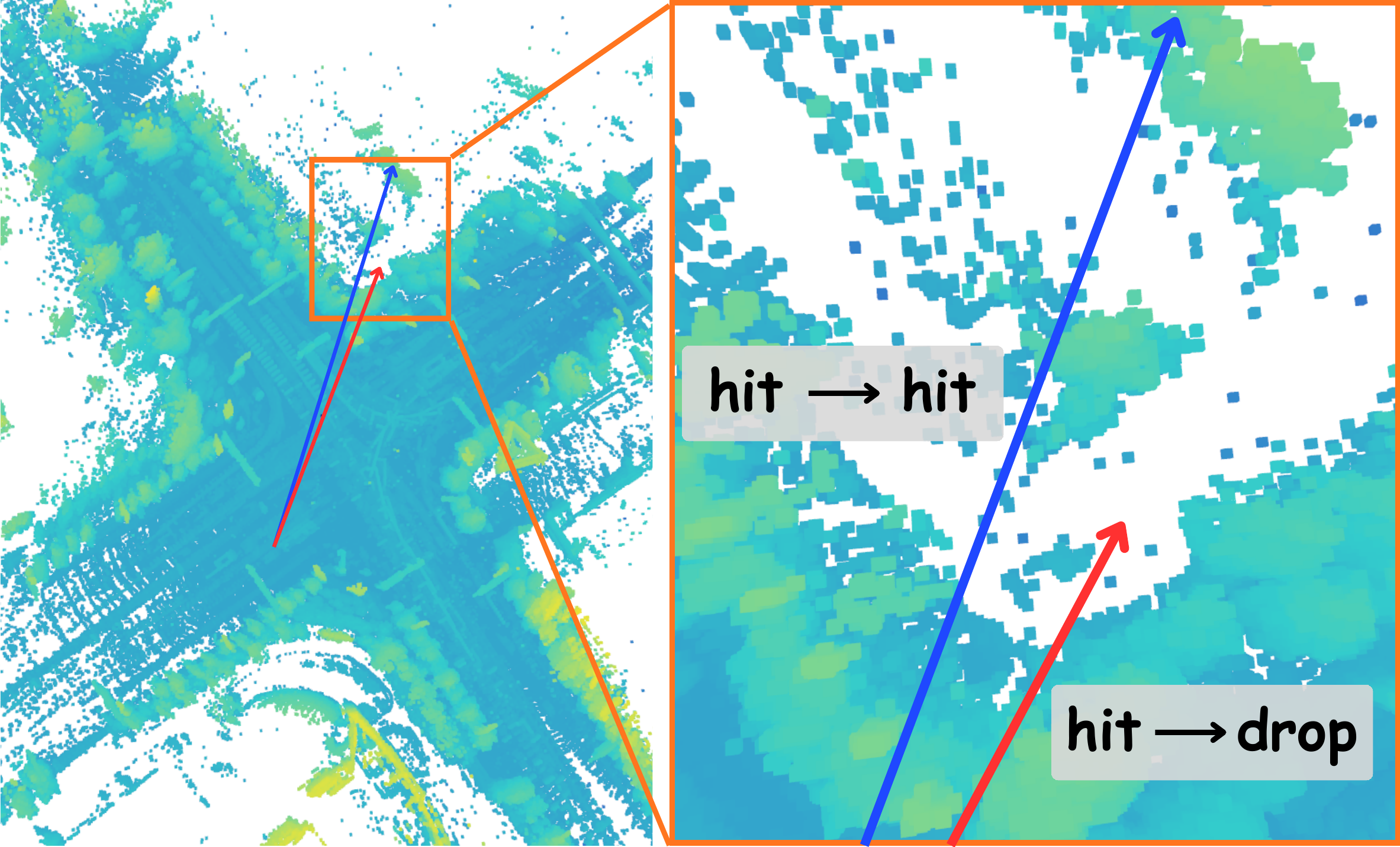}
   \caption{Illustration of the occupancy-based visibility constraint for
background rendering. The blue ray indicates a sample that falls inside the
dilated occupancy region and thus undergoes standard SDF evaluation
(hit → hit), whereas the red ray represents a sample outside the observable
space that is directly classified as drop (hit → drop)}
   \label{fig:pic6}
\end{figure}

To mitigate this extrapolation issue caused by visibility discrepancies, we incorporate a 3D occupancy grid derived from vehicle-side observations into the rendering process. Vehicle-side background point clouds aligned in the world coordinate frame are voxelized into a high-resolution occupancy map representing the spatial extent actually observed during training. Because neural fields tend to extrapolate near the outer fringe of the observable region, directly using the raw occupancy grid may suppress valid structures or truncate portions of the true geometry. To avoid such boundary-induced errors, we apply a morphological dilation to the occupancy grid, introducing a conservative safety margin around the observable space.
During background rendering, each sampled point is checked against the
dilated grid: points inside the occupied region undergo standard SDF and
ray drop evaluation, while points outside are directly classified as drop,
as illustrated in Fig.~\ref{fig:pic6}.
This explicit spatial constraint suppresses unreliable geometry predictions in unobserved regions and prevents background artifacts from occluding or intersecting with foreground vehicles, thereby improving stability under cross-view rendering.

During reconstruction, some vehicles are removed due to insufficient point cloud observations, reducing the number of usable vehicle neural fields from $n$ to $n'$. For trajectories that lack a corresponding neural model, we fill the missing entries by randomly sampling one of the reconstructed $n'$ vehicle neural fields. This strategy ensures that all vehicle trajectories remain valid participants in the rendering pipeline and avoids interruptions caused by missing models.

The overall rendering process follows the neural field formulation of DyNFL \cite{wu2024dynamic} and integrates seamlessly with the above spatial constraints. For each ray, we first determine its intersections with the vehicle bounding boxes in the current frame. If a ray potentially interacts with $k$ vehicles, it is independently evaluated by the corresponding $k$ vehicle neural fields as well as the background neural field, producing $k+1$ candidate LiDAR measurements. For each candidate, we assess its ray drop probability, where background drop prediction follows the occupancy-grid constraint. If all neural fields assign drop probabilities above 0.5, the ray is classified as a drop event. Otherwise, we collect the predicted distances from all valid candidates, sort them in ascending order, and select the smallest distance as the final LiDAR measurement. This selection mechanism preserves correct occlusion relationships between foreground vehicles and the background while ensuring consistent geometry across viewpoints.

 \section{Experiment}

 \subsection{Dataset}
 
To enable direct evaluation of synthesized roadside point clouds against real measurements,
we adopt the V2X-Seq dataset~\cite{yu2023v2x}, which provides synchronized vehicle-side and roadside sensor observations.
The V2X-Seq dataset~\cite{yu2023v2x}
is a large-scale, real-world dataset  for vehicle-infrastructure collaborative perception.  It contains 95 traffic scene sequences collected at six intersections, each lasting 10–20 seconds, with synchronized image and point cloud streams recorded at 10 Hz from both vehicle-side and infrastructure-side sensors.
 The dataset also provides 3D tracking annotations for all objects of interest, where identical objects across frames within a sequence share a unique tracking ID.

 \subsection{Implementation Details}





All experiments were conducted on a workstation equipped with an NVIDIA RTX
4090 GPU, a 24-core Intel CPU, and 64 GB RAM.
 We selected the yizhuang06 intersection  for the experiments to effectively evaluate the proposed method in a controlled environment. We denote the vehicle-side subset as Real-Veh and the corresponding roadside subset as Real-Road.
Using VRS, we synthesize 11196 frames of roadside point clouds from Real-Veh under six custom roadside viewpoints, forming Sim-Road, as illustrated in Fig.~\ref{fig:pdf-example3}.
 To demonstrate the ability of our approach to generate large-scale data
at low cost, we construct a virtual dataset with more synthesized roadside frames than vehicle-side observations.


For 3D object detection, we used PointPillars \cite{lang2019pointpillars} as the primary baseline detector and PV-RCNN \cite{shi2020pv} in some experiments to evaluate the generalizability of the proposed method. All models were trained using their official open-source implementations. The training and testing datasets were split with a 4:1 ratio to ensure fair comparison across different experimental setups. To maintain consistency, the point cloud range was set to \([0, -150.4, -2, 225.6, 150.4, 4]\), with all other hyperparameters kept consistent.


\subsection{Perception Evaluation}
\begin{figure*}[h]
\centering
  \begin{subfigure}[h]{0.235\textwidth}
    \centering
    \includegraphics[width=\textwidth]{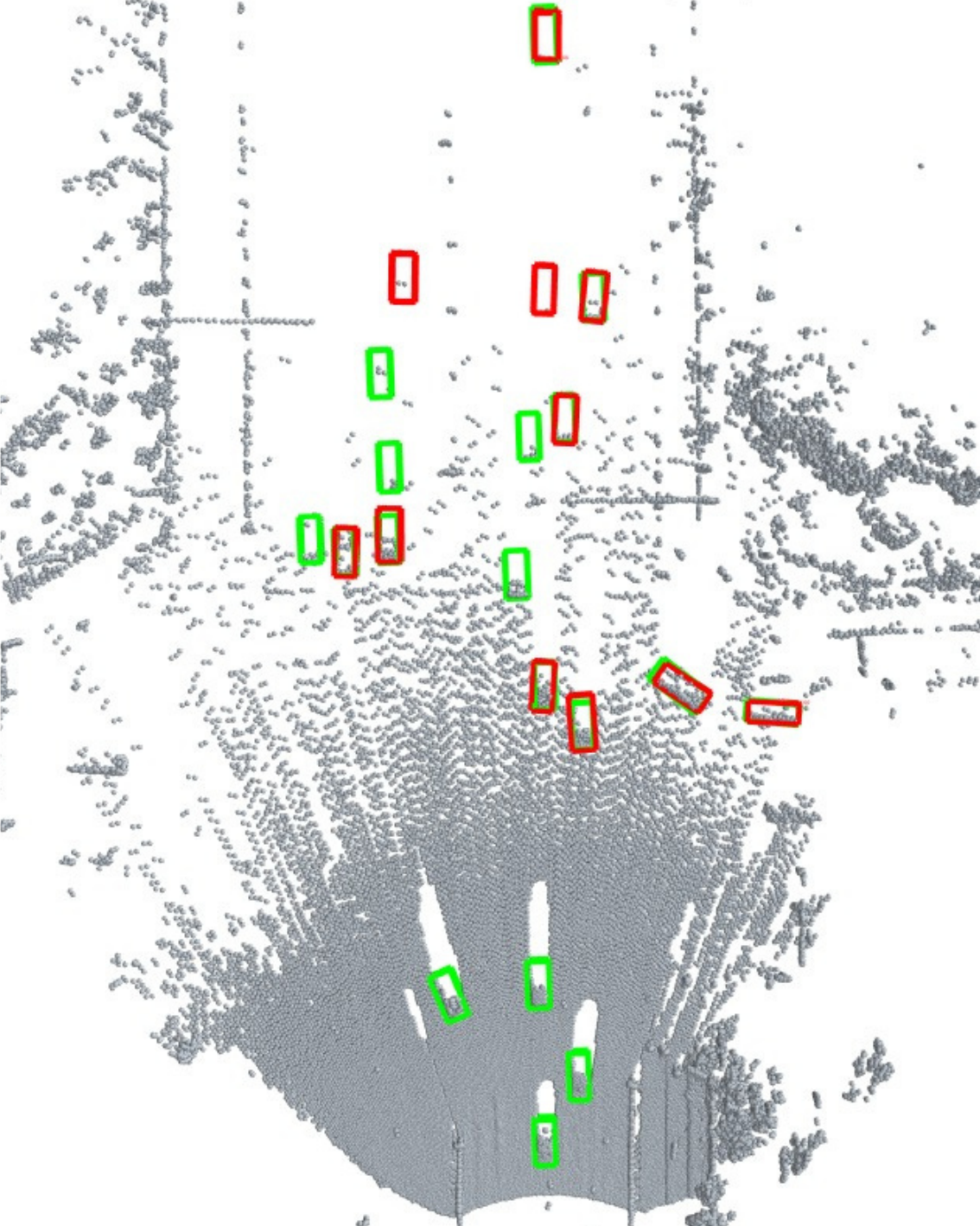}
    \caption*{Real-Veh}
  \end{subfigure}
  \hfill
  \begin{subfigure}[h]{0.235\textwidth}
    \centering
    \includegraphics[width=\textwidth]{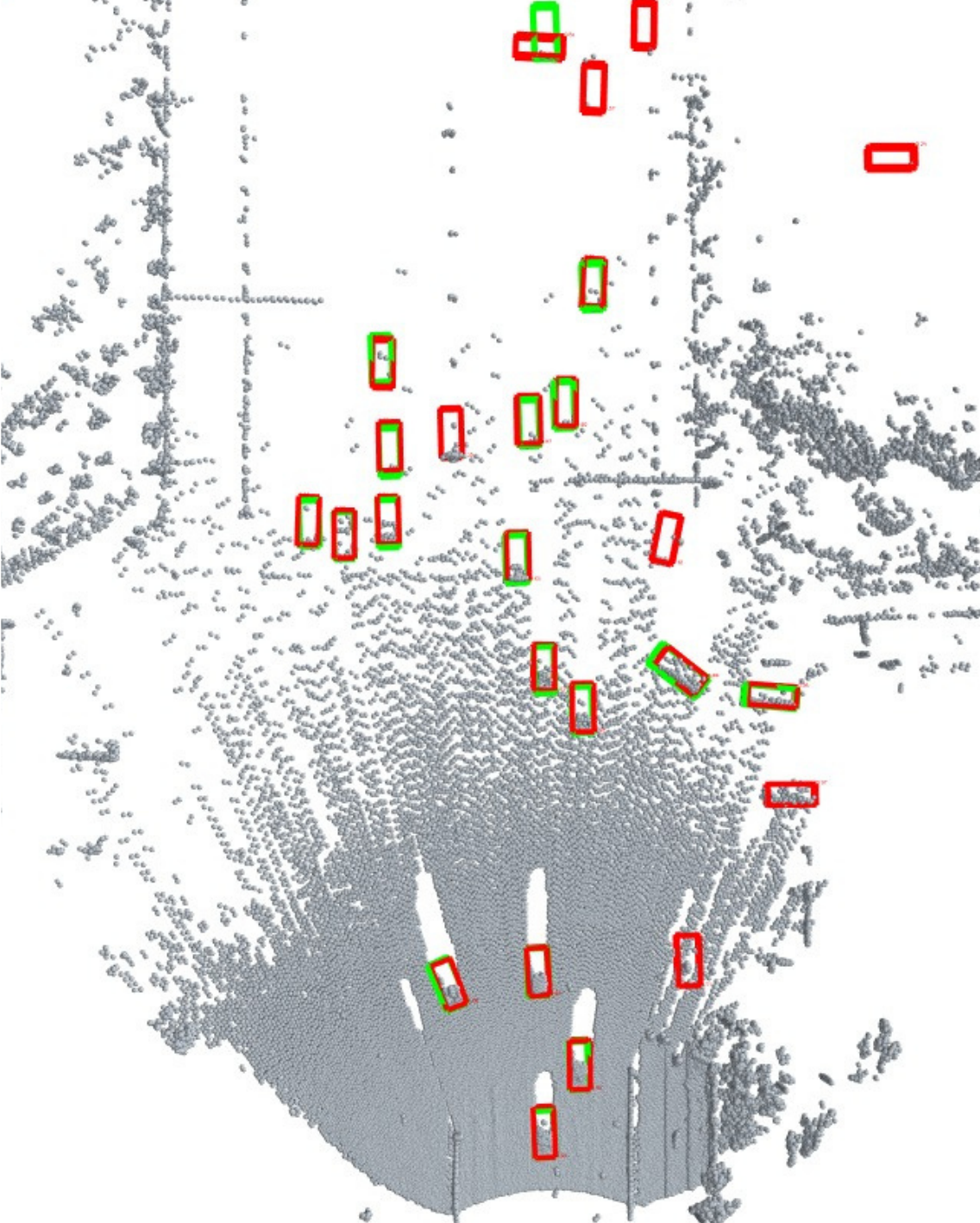}
    \caption*{Sim-Road}
  \end{subfigure}
  \hfill
  \begin{subfigure}[h]{0.235\textwidth}
    \centering
    \includegraphics[width=\textwidth]{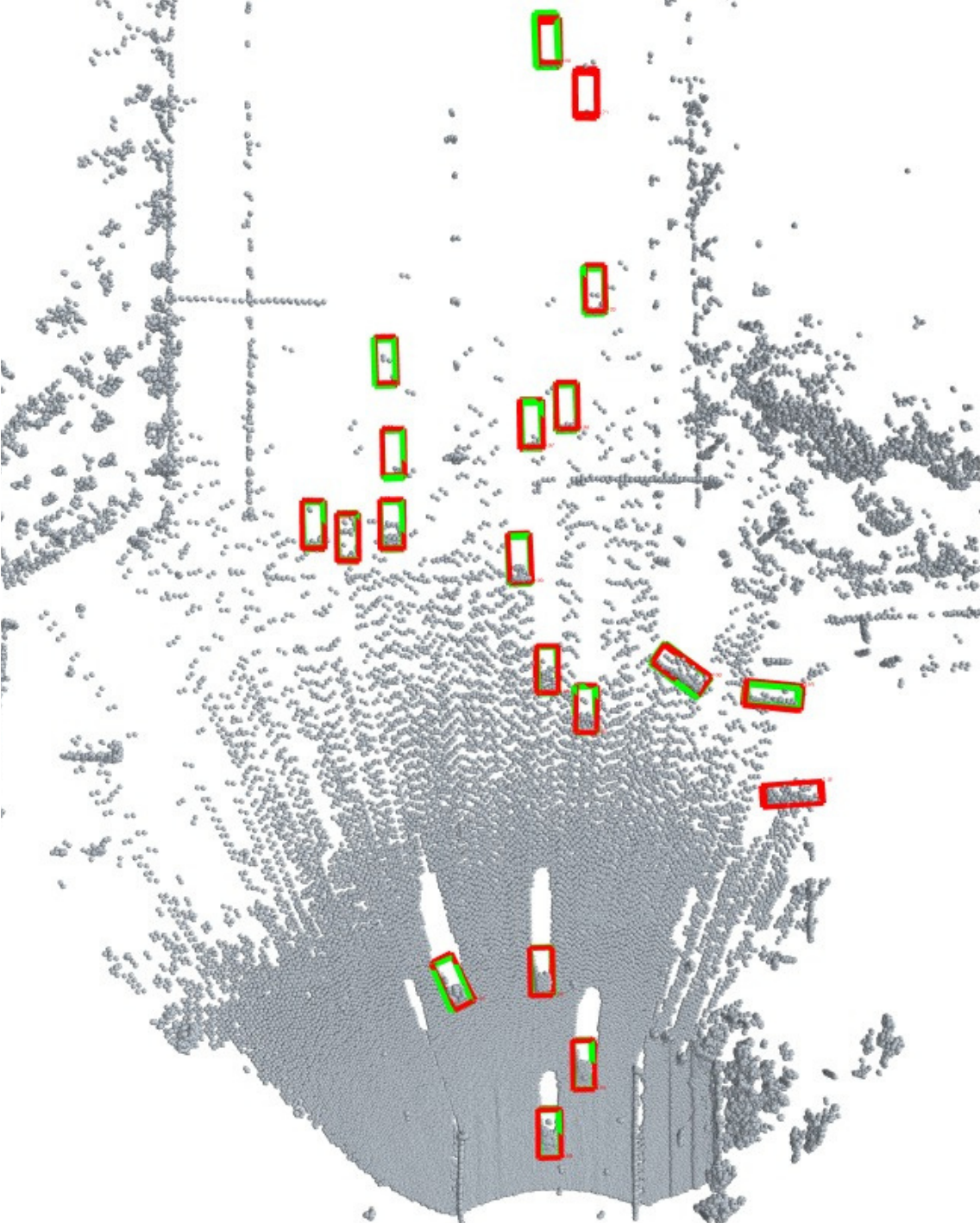}
    \caption*{Real-Road}
  \end{subfigure}
  \hfill
  \begin{subfigure}[h]{0.235\textwidth}
    \centering
    \includegraphics[width=\textwidth]{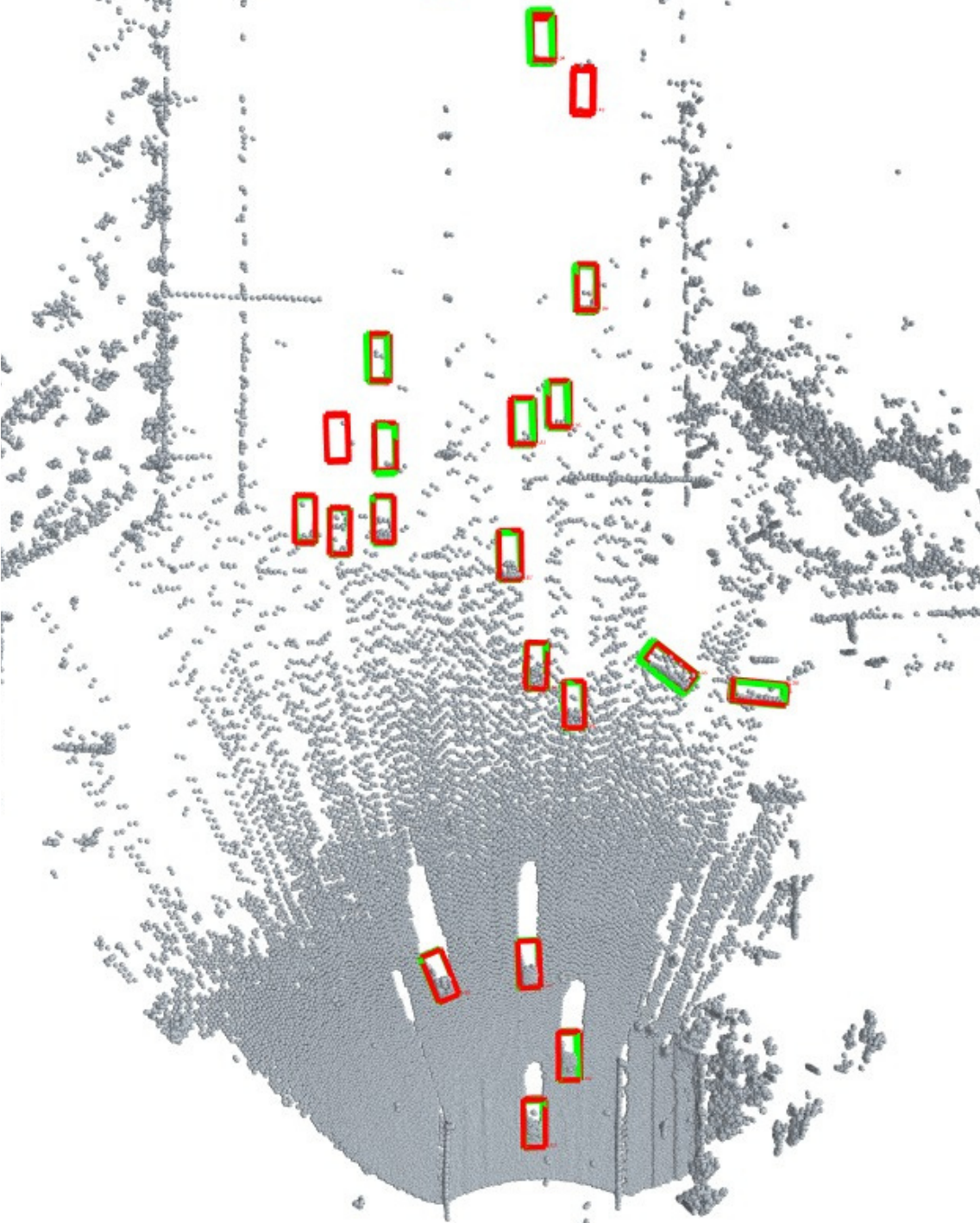}
    \caption*{Real-Road + Sim-Road}
  \end{subfigure}

  \caption{3D object detection results under four different training data configurations.
All models are evaluated on the Real-Road test set. Green and red boxes denote GT and detection respectively.}
  \label{fig:pdf-example}
\end{figure*}

This section evaluates the effectiveness of VRS–based
synthetic roadside data for 3D object detection, with analyses
on overall performance, robustness to limited real roadside
data, generalization to unseen viewpoints, and consistency
across detector architectures.

\subsubsection{Overall Effectiveness}

We first evaluate the overall effectiveness of VRS. Specifically, we train 3D object detection models using four training data settings, namely Real-Veh, Sim-Road, Real-Road, and Real-Road + Sim-Road, and evaluate all models on the test split of Real-Road.


The results are summarized in Table~\ref{tab:overall}. When trained only on
Real-Veh, the model performs poorly on
the roadside test set, achieving only 8.9\% 3D AP@0.7,
which reveals a substantial viewpoint and point-cloud
distribution discrepancy between vehicle-side and roadside
sensing. 
In contrast, training exclusively on Sim-Road improves the
performance to 42.2\% 3D AP@0.7, corresponding to an almost five-fold improvement
over vehicle-side training. 
This improvement indicates that VRS effectively alleviates the vehicle-to-roadside domain discrepancy, while the strong detection performance also suggests that the synthesized roadside point clouds preserve meaningful geometric structures for downstream 3D object detection.
Training on Real-Road achieves
66.1\% 3D AP@0.7, and further incorporating synthetic
roadside LiDAR boosts the performance to 72.1\%
(+6.0 AP), with consistent gains also observed at IoU 0.5.
These results clearly indicate that VRS
provides effective complementary supervision to real roadside
data.

\subsubsection{Effect under Limited Real-World Data}

In practice, large-scale real-world roadside LiDAR data is
difficult to collect and often limited in both spatial coverage
and temporal diversity. To reflect this constraint, we evaluate
the effectiveness of synthetic roadside data under different
amounts of real roadside training data(Road) by restricting the
training set to 10\%, 30\%, 50\%, and
100\%, while keeping all other settings unchanged.

As shown in Fig.~\ref{fig:pdf-example}, incorporating VRS consistently
improves detection performance across all data ratios, with
particularly pronounced gains when real roadside data is
limited. When only 10\% of the real data is available,
the baseline achieves 12.1\% 3D AP@0.7, while adding
synthetic roadside data boosts the performance to
49.3\%, yielding a gain of +37.3 AP. As the
amount of real data increases, the absolute gains gradually
decrease but remain substantial, and consistent improvements
are still observed even under full-data training. These results
demonstrate that VRS effectively mitigates the
limitations of real-world roadside data availability by
providing scalable and informative synthetic supervision.
\begin{figure}[htbp]
   \centering
   \includegraphics[width=0.43\textwidth]{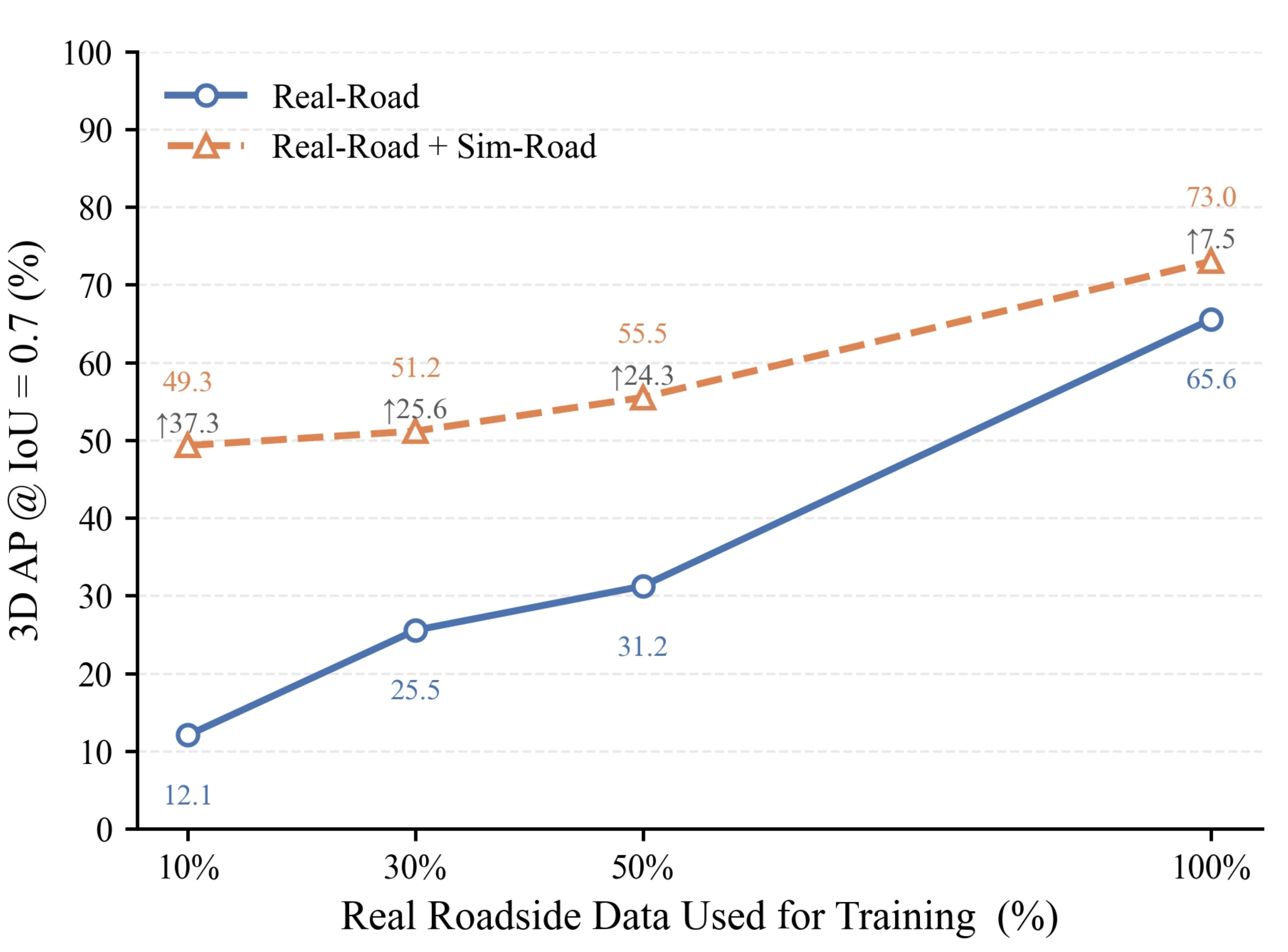}
   \caption{Effect of VRS synthetic data under different real roadside data budgets.}
   \label{fig:pdf-example}
\end{figure}

\subsubsection{Viewpoint Generalization}

 We evaluate the impact of synthetic roadside data generated by VRS on the viewpoint generalization of 3D object detection through data augmentation. To this end, we train detection models using 100\% real roadside data (Road), with or without augmentation by synthetic roadside point clouds (Sim-Road). The trained models are then directly evaluated on an unseen-viewpoint test set constructed by randomly sampling 1,000 frames from five other intersections in V2X-Seq-I (the roadside subset of V2X-Seq), that are not used during training.

As shown in Table~\ref{tab:unseen}, detection performance drops significantly
when evaluated on unseen viewpoints. Training on Road
achieves 38.1\% 3D AP@0.7 on the unseen-viewpoint test
set, compared to 66.1\% on same-intersection tests.
Adding synthetic roadside data from Sim-Road improves the
performance to 46.7\% 3D AP@0.7(+8.6 AP), indicating that
VRS improves viewpoint generalization by expanding the viewpoint diversity of the training data.

\subsubsection{Model Generality}

To evaluate detector generality, we conduct supplementary
experiments using PV-RCNN \cite{shi2020pv}. 
The amount of real roadside training data from Road is fixed to 50\%,
representing a limited yet non-trivial data regime.
Models are trained with and without data augmentation using synthetic roadside data from Sim-Road under the same configurations as in previous experiments.

As shown in Table~\ref{tab:pvrcnn}, PV-RCNN \cite{shi2020pv} achieves 46.3\% 3D
AP@0.7 when trained on 50\% of the real roadside data
from Road. Applying data augmentation with synthetic
roadside data improves the performance to 53.3\%
(+7.0 AP), with consistent gains also observed at IoU 0.5.
This trend is consistent with the results obtained using
PointPillars \cite{lang2019pointpillars}, indicating that the synthetic roadside data obtained using
VRS enables effective data augmentation across
different detector architectures.

\begin{table}[t]
\centering
\caption{ Overall performance of roadside 3D object detection on real roadside
data under 4 different training data configurations
}
\label{tab:overall}
\small
\setlength{\tabcolsep}{4pt}
\renewcommand{\arraystretch}{1.15}
\begin{tabular*}{\columnwidth}{@{\extracolsep{\fill}}lcccc}
\hline
Training Data 
& \multicolumn{2}{c}{IoU = 0.7} 
& \multicolumn{2}{c}{IoU = 0.5} \\
& AP$_{\text{BEV}}$ & AP$_{\text{3D}}$
& AP$_{\text{BEV}}$ & AP$_{\text{3D}}$ \\
\hline
Real-Veh & 45.55 & 8.91 & 55.43 & 44.05 \\
Sim-Road & 79.15 & 42.18 & 86.15 & 81.74 \\
\hline
Real-Road & 91.55 & 66.14 & 97.45 & 91.95 \\
\makecell[l]{Real-Road + Sim-Road}
& \makecell{\textbf{93.00}\\(+1.45)}
& \makecell{\textbf{72.14}\\(+6.00)}
& \makecell{\textbf{98.24}\\(+0.79)}
& \makecell{\textbf{95.51}\\(+3.56)} \\
\hline
\end{tabular*}
\end{table}




\begin{table}[t]
\centering
\caption{Roadside 3D object detection performance on unseen intersections for viewpoint generalization evaluation}
\label{tab:unseen}
\small
\setlength{\tabcolsep}{4pt}
\renewcommand{\arraystretch}{1.15}
\begin{tabular*}{\columnwidth}{@{\extracolsep{\fill}}lcccc}
\hline
Training Data 
& \multicolumn{2}{c}{IoU = 0.7} 
& \multicolumn{2}{c}{IoU = 0.5} \\
& AP$_{\text{BEV}}$ & AP$_{\text{3D}}$
& AP$_{\text{BEV}}$ & AP$_{\text{3D}}$ \\
\hline
Real-Road & 60.86 & 38.12 & 66.75 & 62.18 \\
\makecell[l]{Real-Road + Sim-Road}
& \makecell{\textbf{66.96}\\(+6.10)}
& \makecell{\textbf{46.69}\\(+8.57)}
& \makecell{\textbf{72.91}\\(+6.16)}
& \makecell{\textbf{69.26}\\(+7.08)} \\
\hline
\end{tabular*}
\end{table}





\begin{table}[h]
\centering
\caption{Detector generality evaluation using PV-RCNN with limited real roadside training data}
\label{tab:pvrcnn}
\small
\setlength{\tabcolsep}{4pt}
\renewcommand{\arraystretch}{1.15}
\begin{tabular*}{\columnwidth}{@{\extracolsep{\fill}}lcccc}
\hline
Training Data 
& \multicolumn{2}{c}{IoU = 0.7} 
& \multicolumn{2}{c}{IoU = 0.5} \\
& AP$_{\text{BEV}}$ & AP$_{\text{3D}}$
& AP$_{\text{BEV}}$ & AP$_{\text{3D}}$ \\
\hline
\makecell[l]{50\% Real-Road} & 59.09 & 46.34 & 64.91 & 62.24 \\
\makecell[l]{50\% Real-Road \\+ Sim-Road\\}
& \makecell{\textbf{73.63}\\(+14.54)}
& \makecell{\textbf{53.26}\\(+6.92)}
& \makecell{\textbf{78.82}\\(+13.91)}
& \makecell{\textbf{76.08}\\(+13.84)} \\
\hline
\end{tabular*}
\end{table}

 \subsection{Ablation Study}

\begin{figure}[!t]
  \centering

  \begin{subfigure}[t]{0.49\columnwidth}
    \centering
    \includegraphics[width=\linewidth]{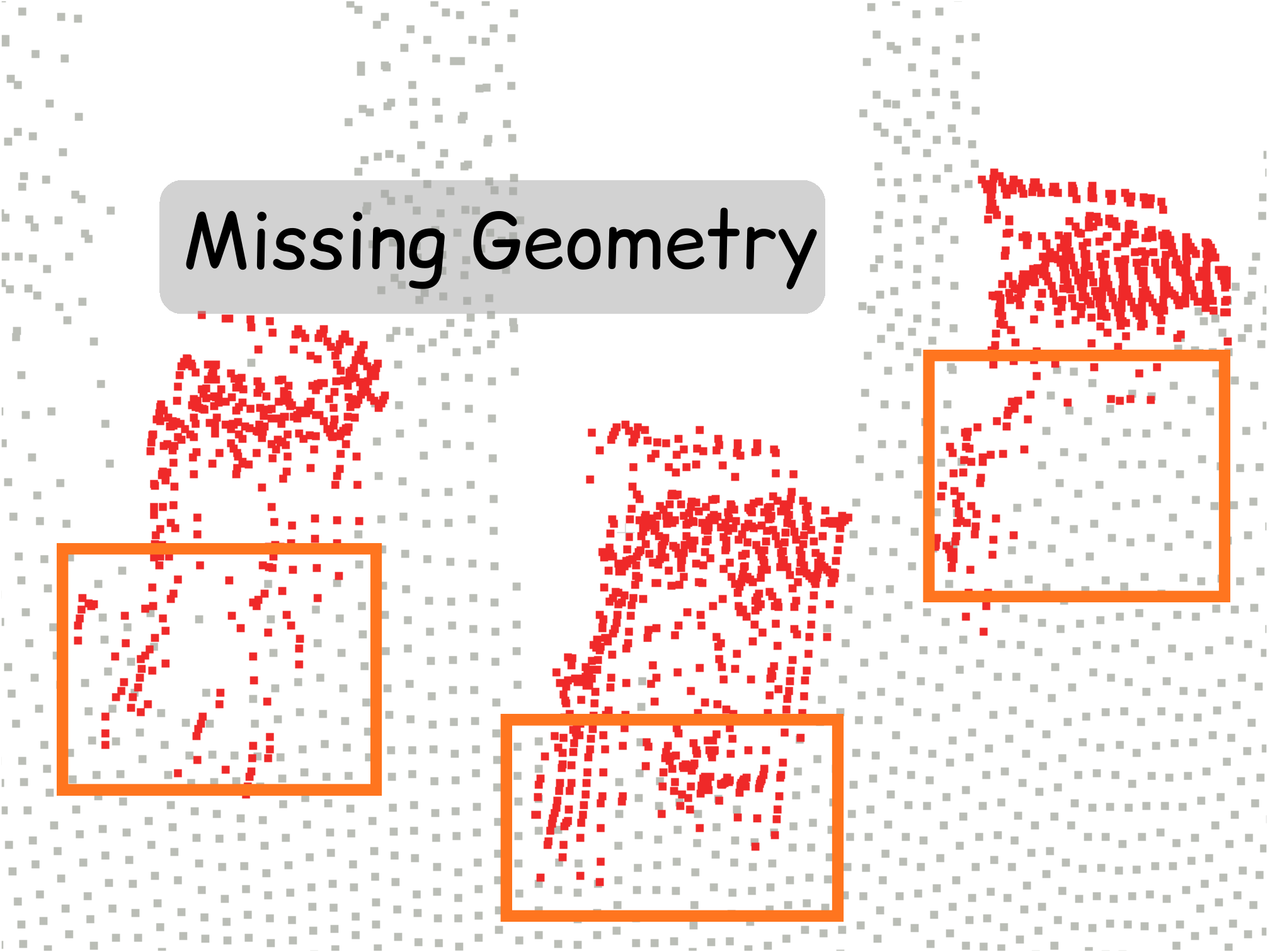}
    \caption*{w/o Point Cloud Completion}
  \end{subfigure}\hfill
  \begin{subfigure}[t]{0.49\columnwidth}
    \centering
    \includegraphics[width=\linewidth]{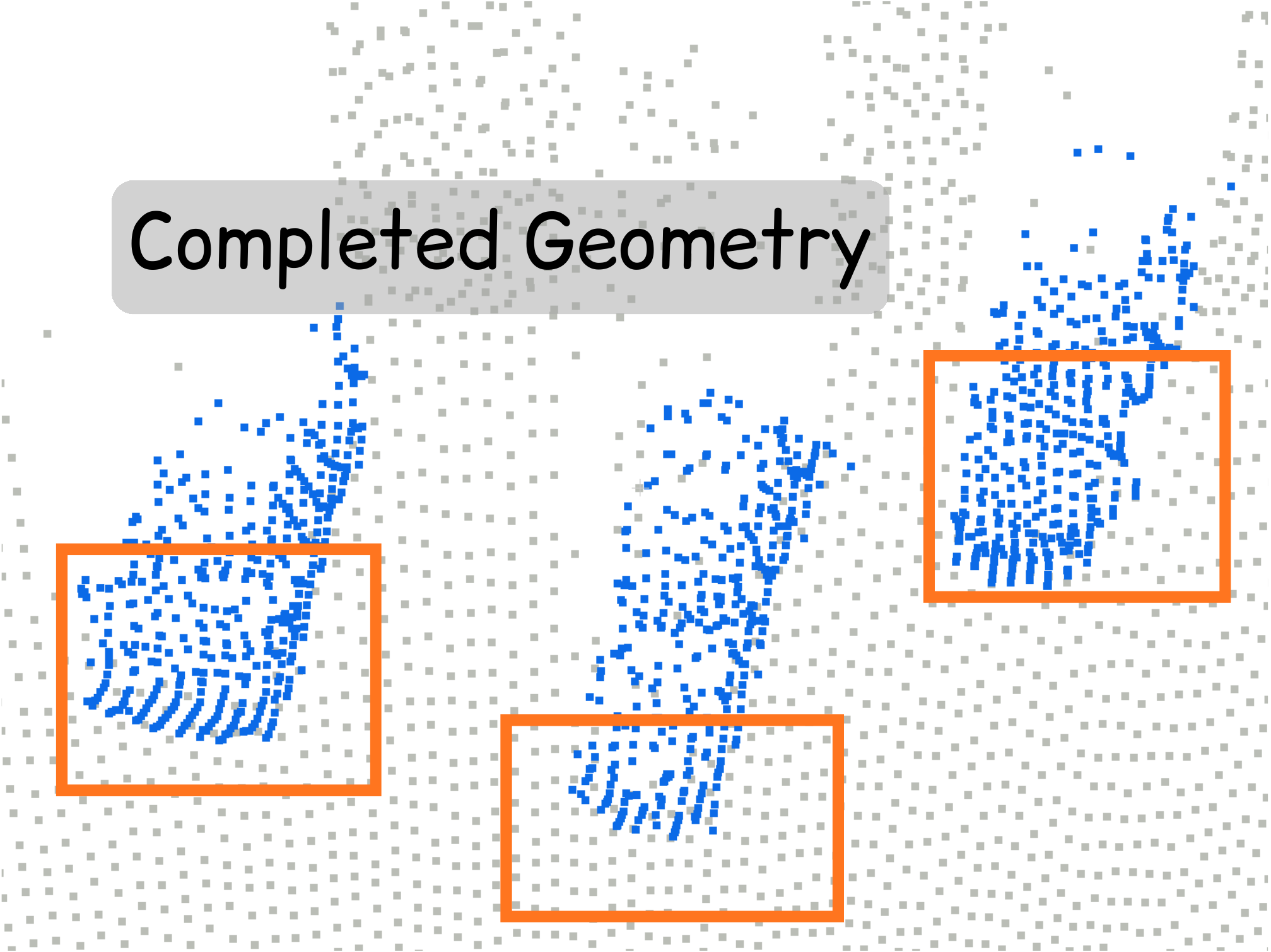}
    \caption*{w. Point Cloud Completion}
  \end{subfigure}

  \vspace{1mm}

  \begin{subfigure}[t]{0.49\columnwidth}
    \centering
    \includegraphics[width=\linewidth]{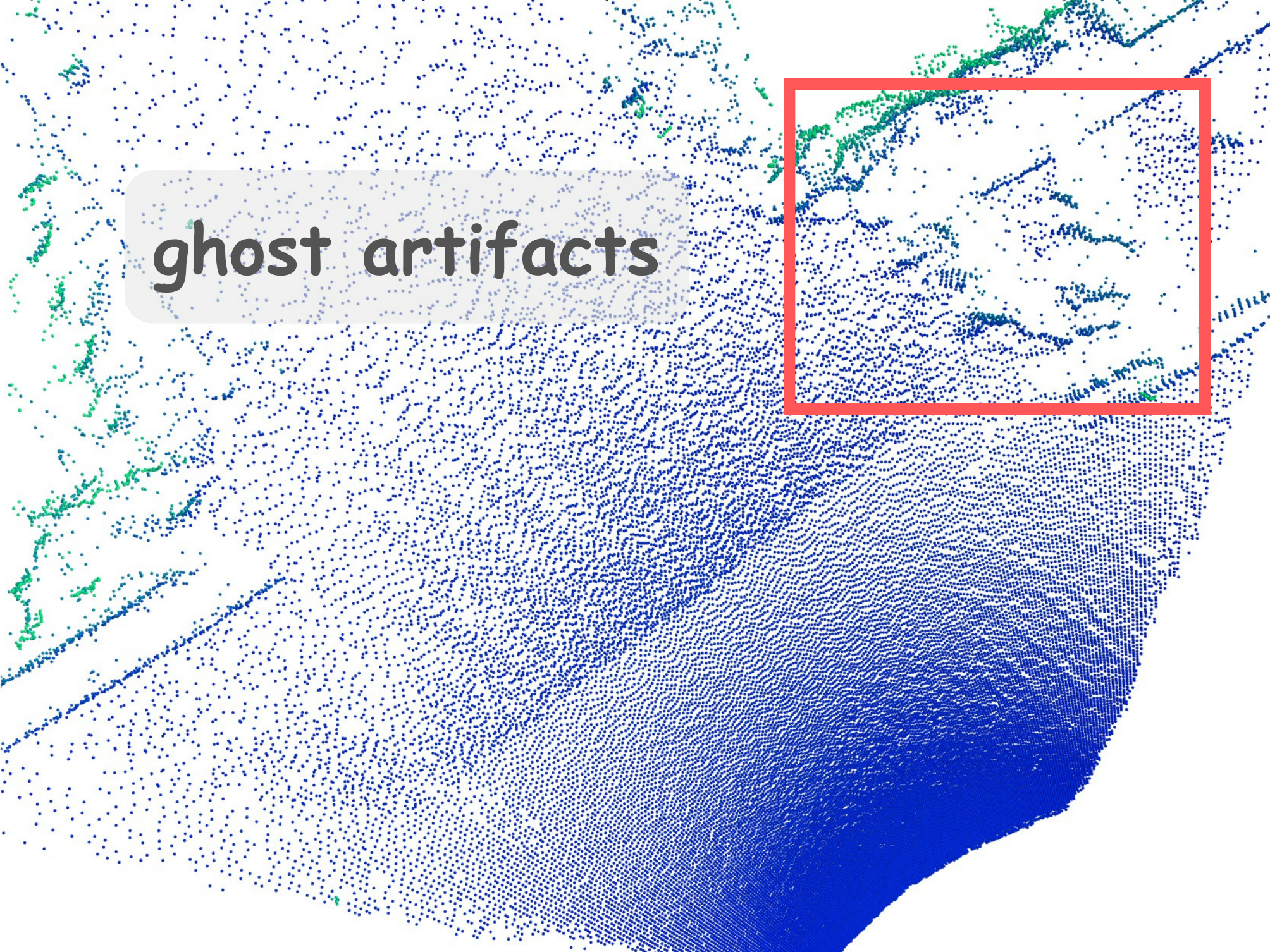}
    \caption*{w/o Dynamic Object Removal}
  \end{subfigure}\hfill
  \begin{subfigure}[t]{0.49\columnwidth}
    \centering
    \includegraphics[width=\linewidth]{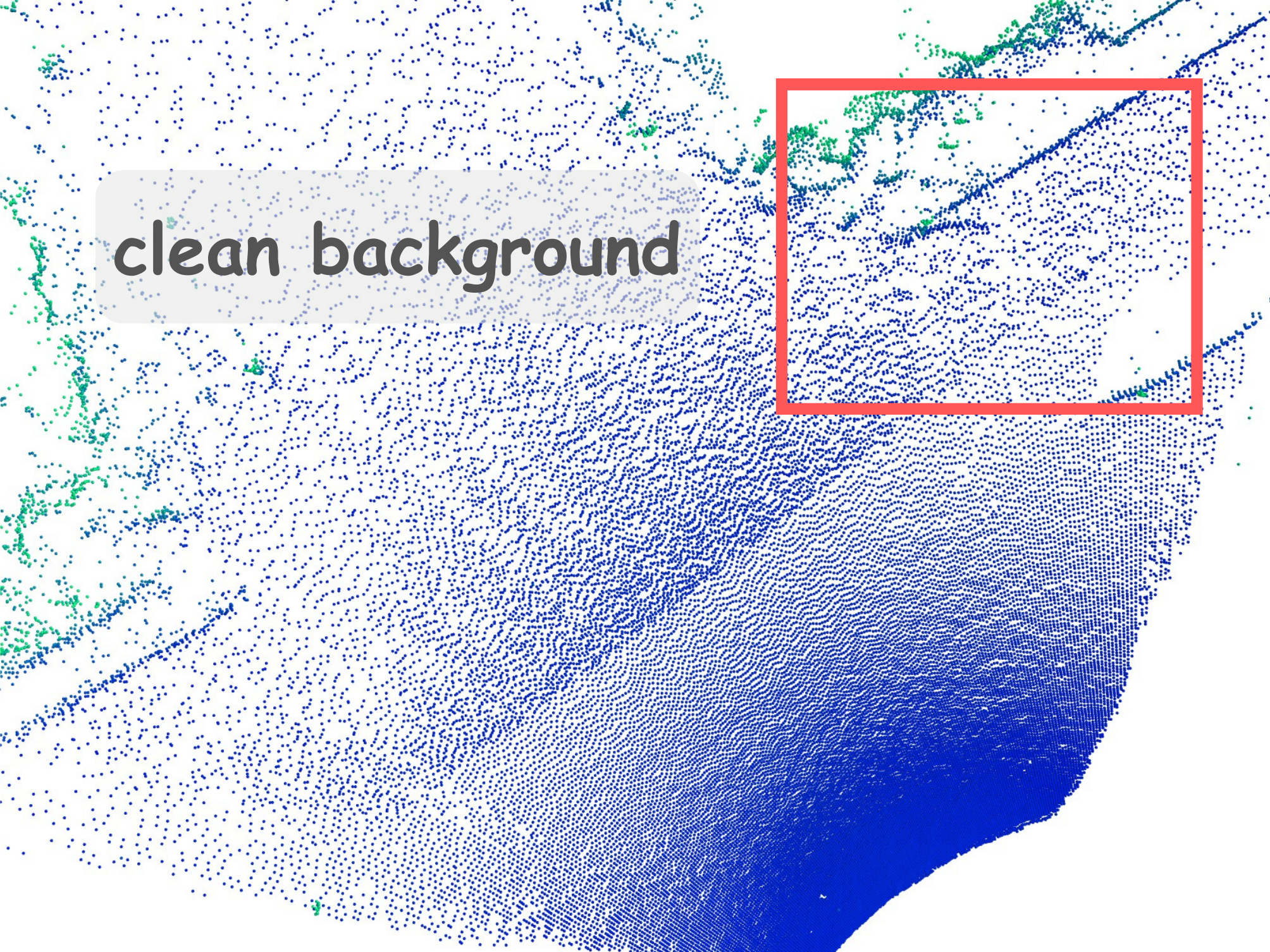}
    \caption*{w. Dynamic Object Removal}
  \end{subfigure}

  \vspace{1mm}

  \begin{subfigure}[t]{0.49\columnwidth}
    \centering
    \includegraphics[width=\linewidth]{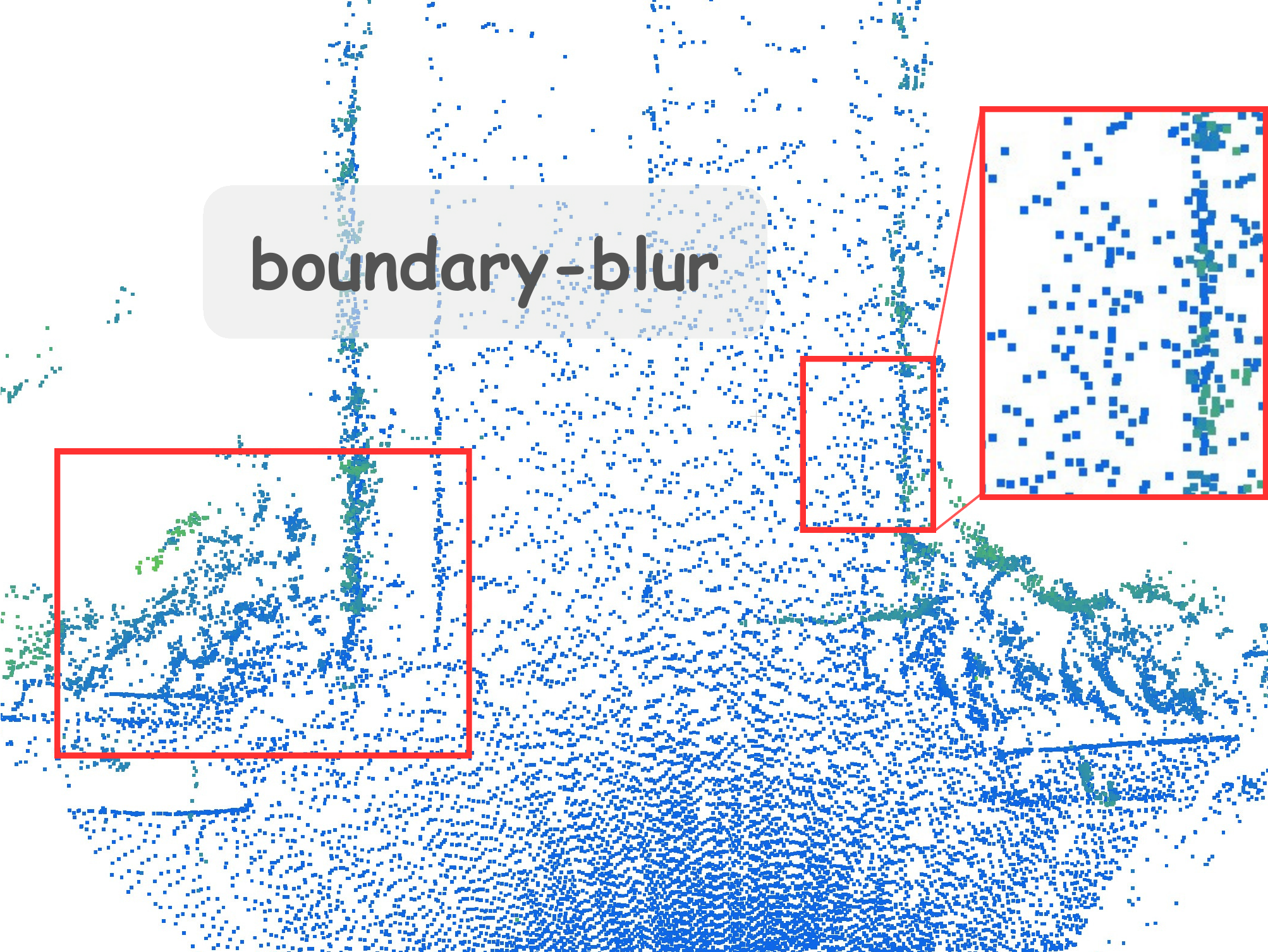}
    \caption*{w/o Global Pose Alignment}
  \end{subfigure}\hfill
  \begin{subfigure}[t]{0.49\columnwidth}
    \centering
    \includegraphics[width=\linewidth]{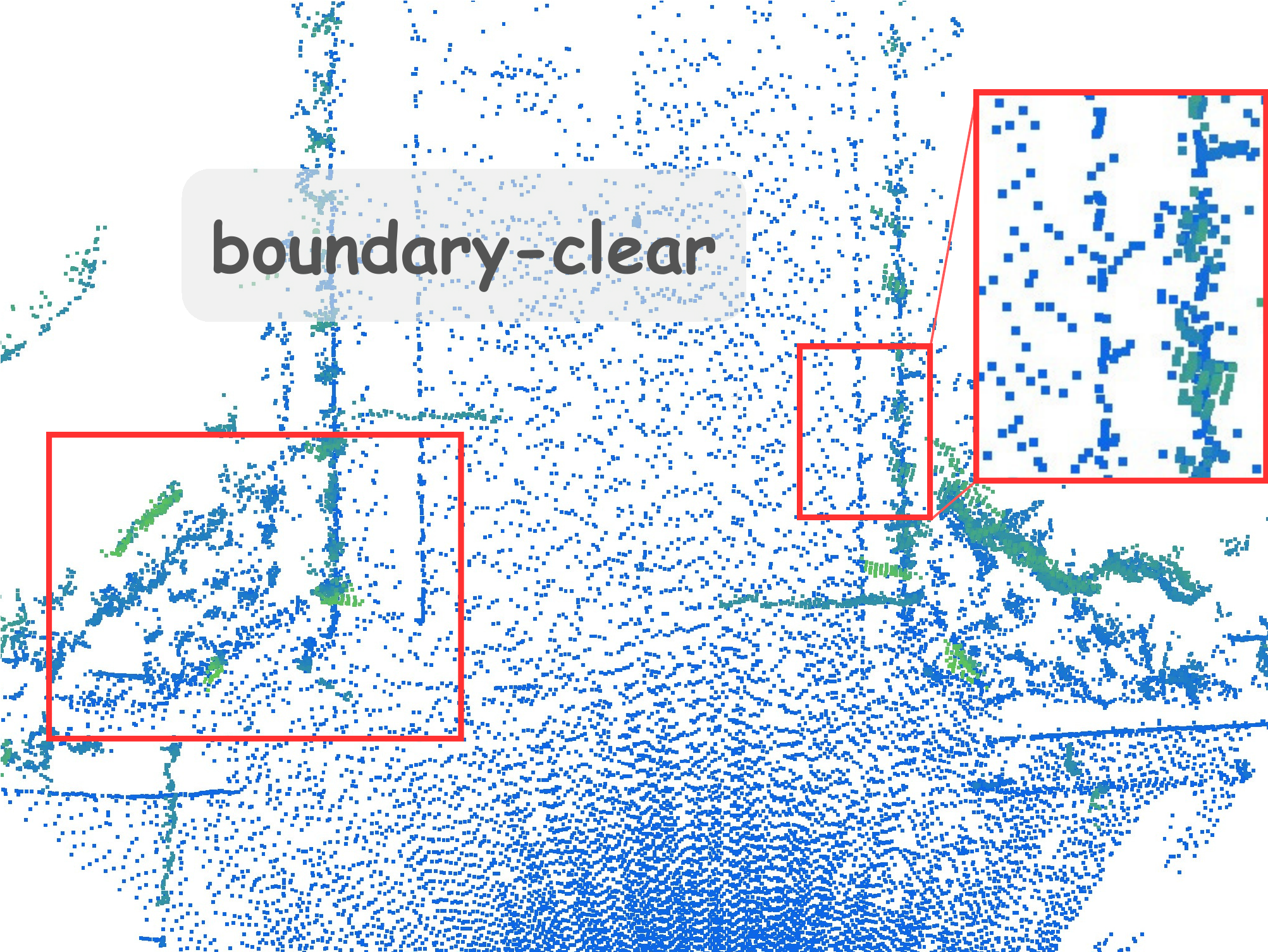}
    \caption*{w. Global Pose Alignment}
  \end{subfigure}

  \vspace{1mm}

  \begin{subfigure}[t]{0.49\columnwidth}
    \centering
    \includegraphics[width=\linewidth]{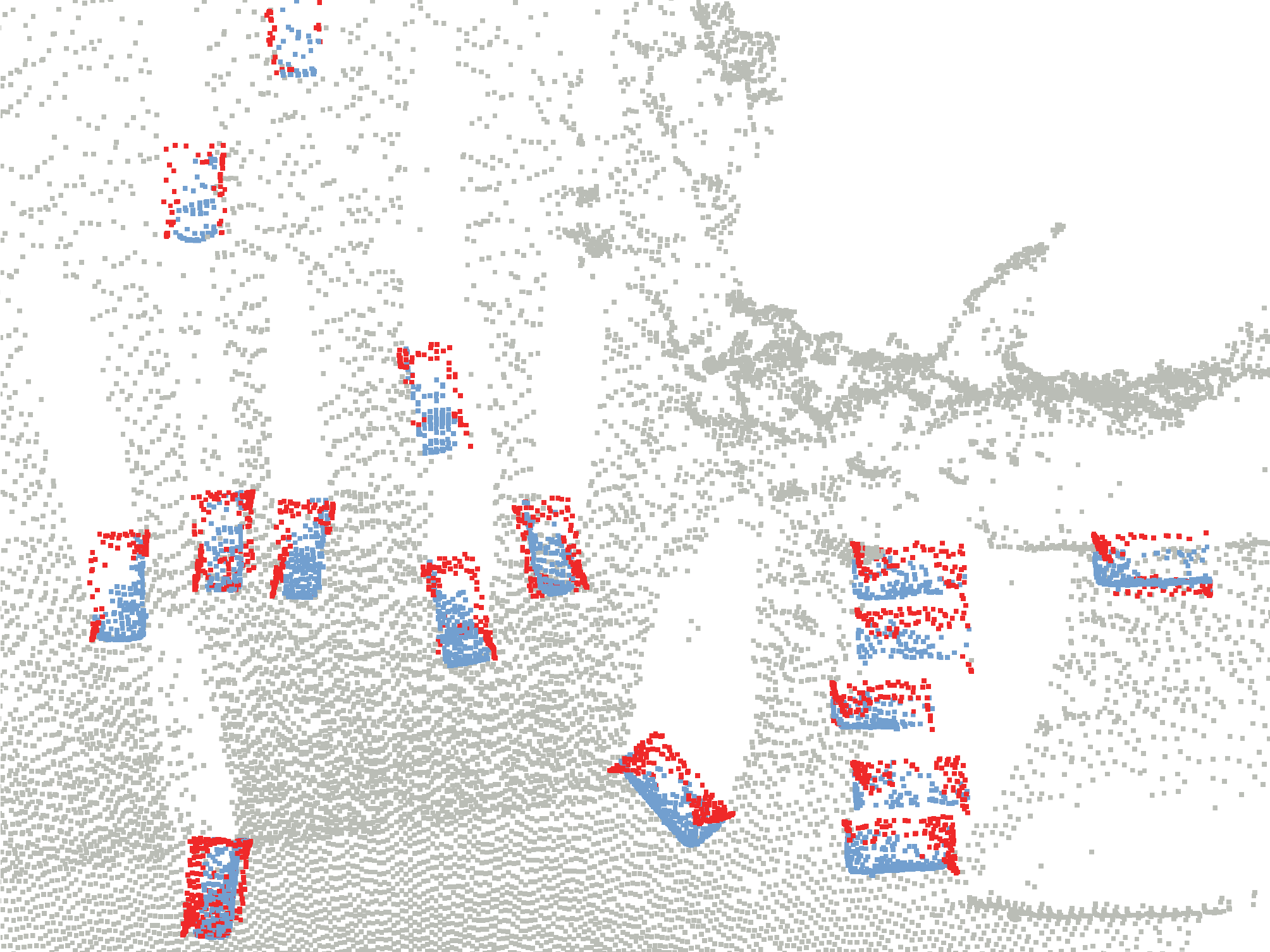}
    \caption*{w/o Ray Drop}
  \end{subfigure}\hfill
  \begin{subfigure}[t]{0.49\columnwidth}
    \centering
    \includegraphics[width=\linewidth]{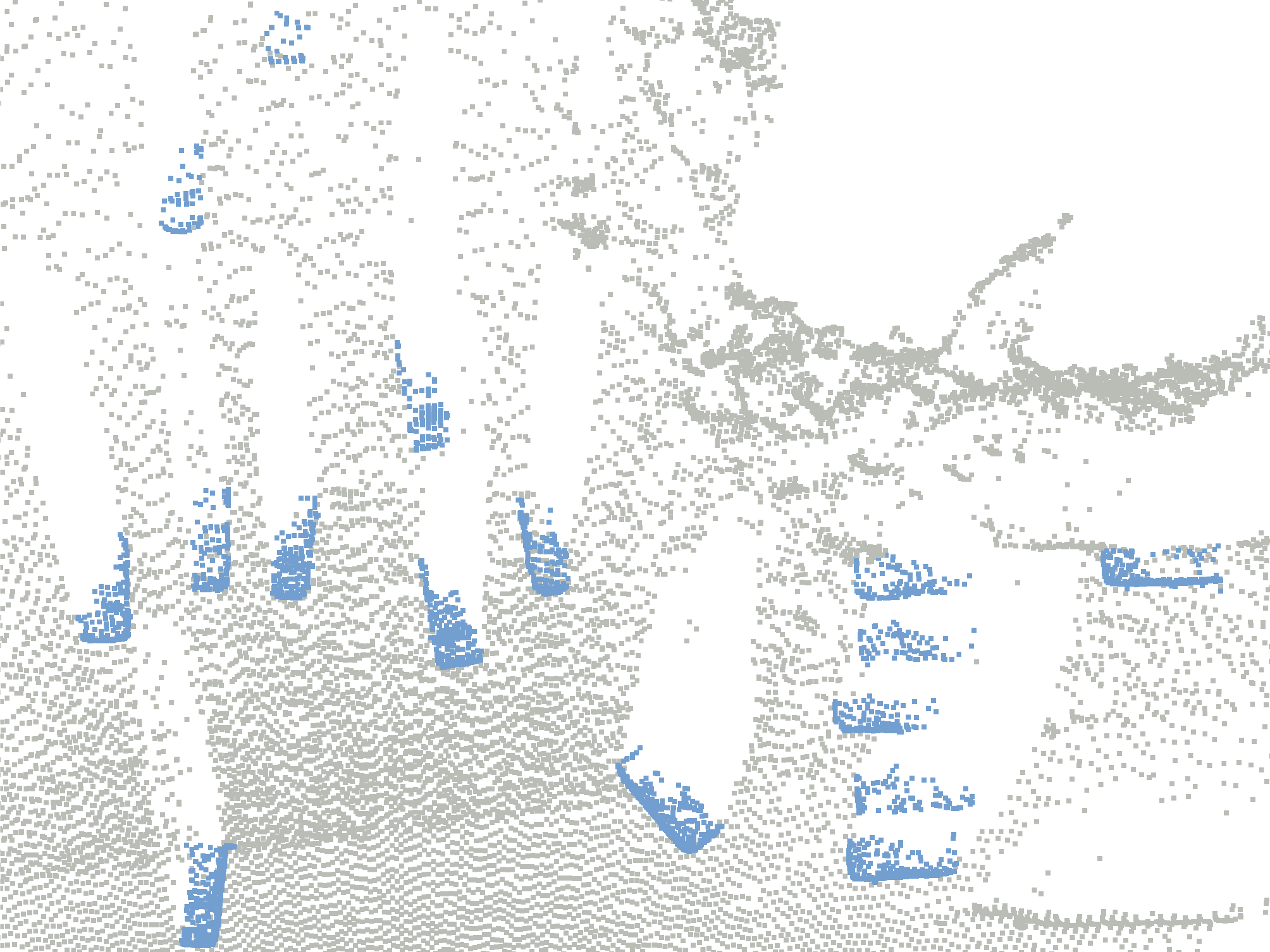}
    \caption*{w. Ray Drop}
  \end{subfigure}

  \vspace{1mm}

  \begin{subfigure}[t]{0.49\columnwidth}
    \centering
    \includegraphics[width=\linewidth]{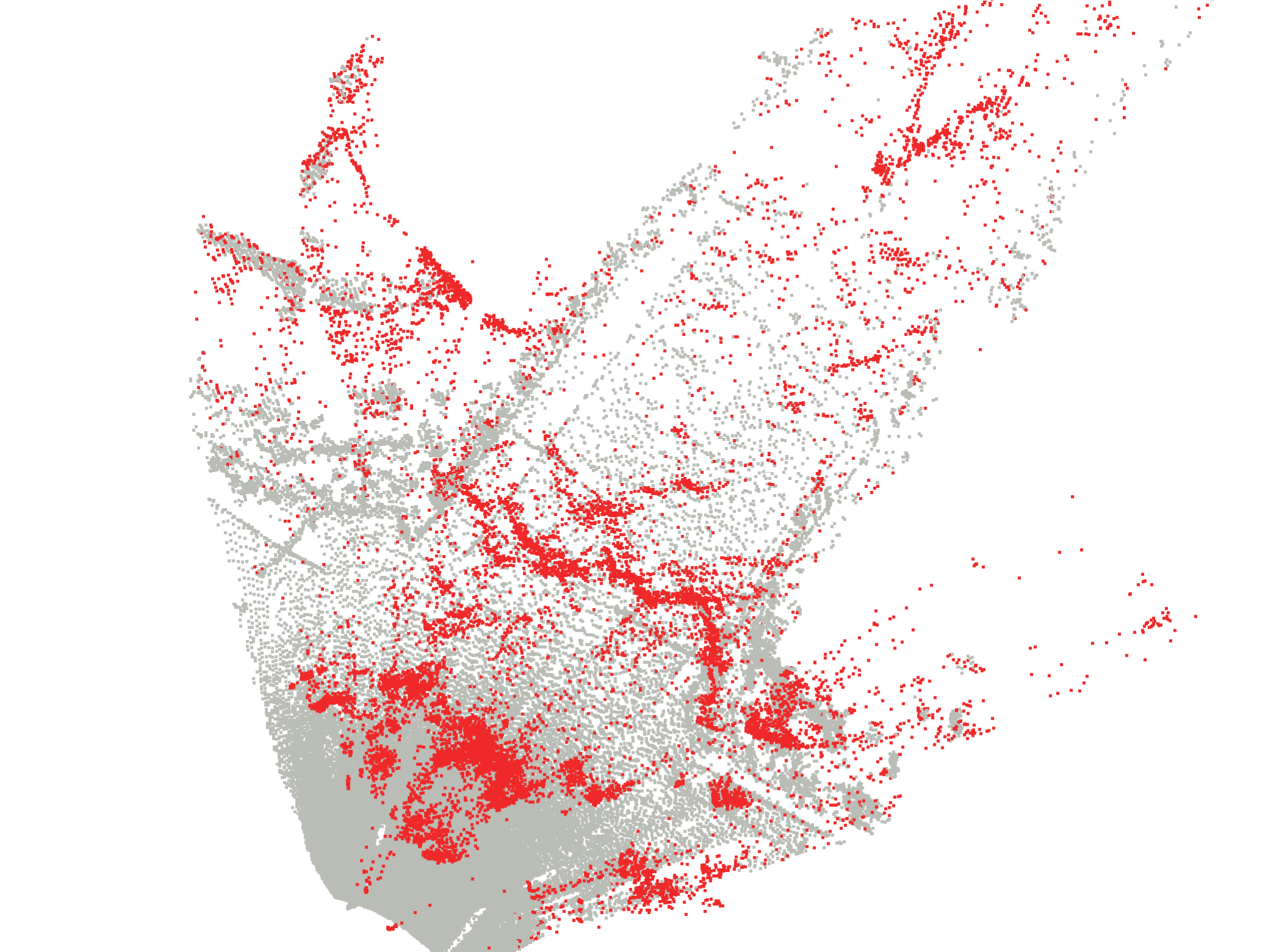}
    \caption*{w/o Occupancy Vis.}
  \end{subfigure}\hfill
  \begin{subfigure}[t]{0.49\columnwidth}
    \centering
    \includegraphics[width=\linewidth]{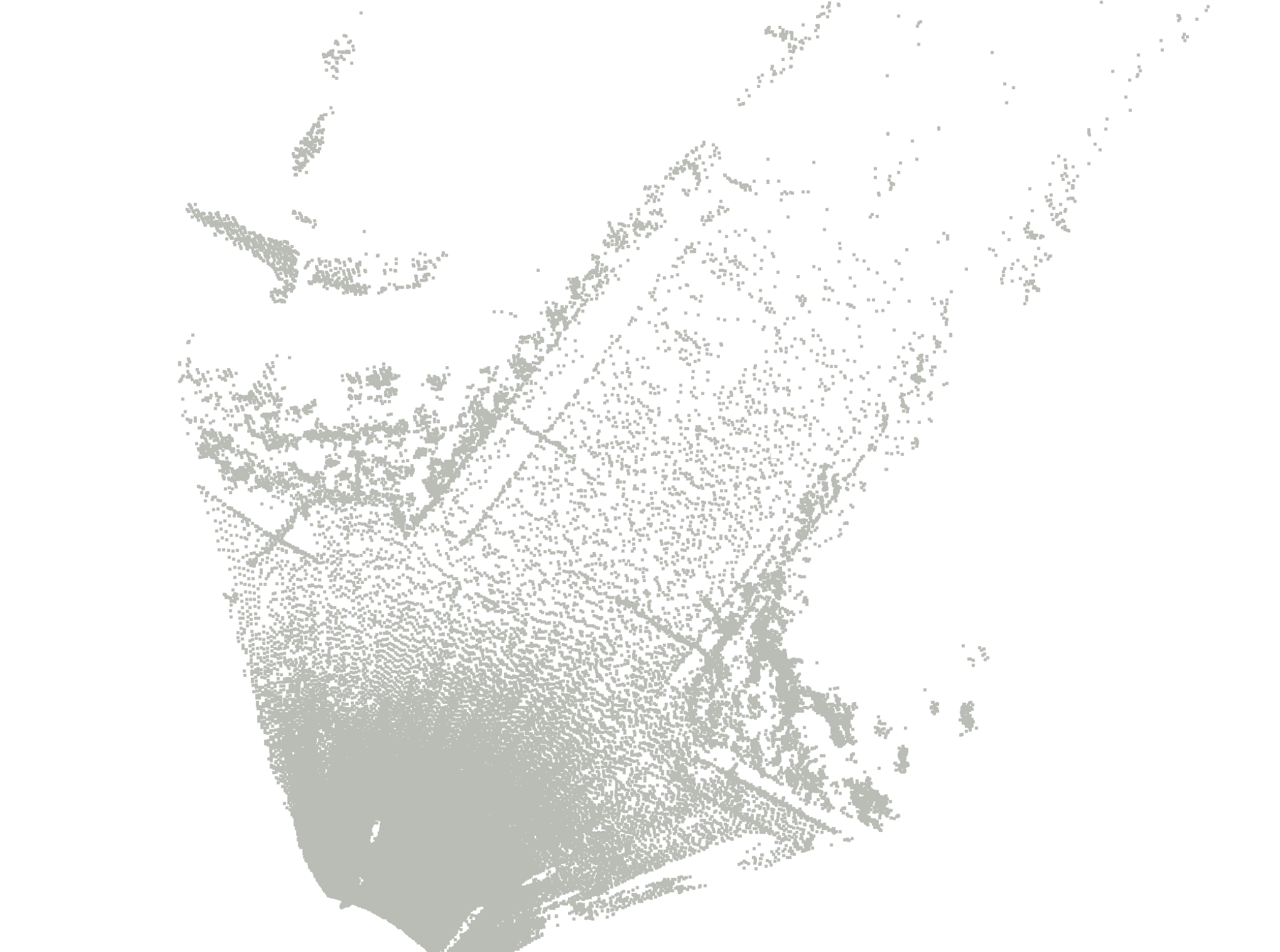}
    \caption*{w. Occupancy Vis.}
  \end{subfigure}

  \vspace{1mm}

  \caption{Qualitative ablation results of key modules.}
  \label{fig:ablation_qual}

\end{figure}

We verify the key designs of VRS, as illustrated in Fig.~\ref{fig:ablation_qual}.

 \textbf{Vehicle Point Cloud Completion:}  The vehicle point cloud completion module solves the problem of missing vehicle geometry due to the limitations of the vehicle-side viewpoint.  This module restores the vehicle's roof and rear structures, which are heavily occluded at the vehicle-side, ensuring that the reconstructed vehicle point clouds have a complete shape from the roadside viewpoint.

 \textbf{Background Dynamic Object Removal:}  Background dynamic object removal eliminates unannotated dynamic vehicles that would otherwise be treated as static structures during background reconstruction. Without this step, dynamic objects contaminate the static background field and lead to ghost artifacts.

\textbf{Global Pose Alignment:} Global pose alignment ensures pose consistency across multiple vehicle-side sequences and improves the geometric quality of the reconstructed background. Without pose optimization, pose errors introduce inconsistent ray constraints, causing the neural field to blur the geometry and fail to capture fine structural details.

 \textbf{Ray Drop Test:} We introduce ray drop probability modeling based on the neural field.  By jointly optimizing surface geometry and ray drop probability, our method learns a more complete LiDAR scan measurement generation process, ensuring that point clouds are generated only at intersections with real surfaces during dynamic vehicle reconstruction.  This significantly improves the physical plausibility of the synthetic data.

 \textbf{Occupancy-based Visibility Constraint:} 
The occupancy-based visibility constraint restricts neural field evaluation to spatial regions supported by vehicle-side observations. Without this constraint, unconstrained extrapolation in unobserved regions leads to spurious background artifacts under roadside viewpoints.

\section{CONCLUSIONS}

This paper presents  Vehicle-to-Roadside LiDAR Synthesis (VRS), a novel framework that synthesizes labeled roadside LiDAR data from vehicle-side datasets to address the scarcity of real-world roadside datasets. Our approach mitigates the vehicle-to-roadside domain gap through point cloud completion and occupancy-based visibility constraint. Experimental results demonstrate that the synthesized data effectively complements real roadside datasets, mitigates the limitations of data scarcity, and improves generalization to unseen roadside viewpoints.

Despite these advancements, VRS currently focuses on car objects and does not address more challenging categories such as cyclists and pedestrians. 
Additionally, while achieving geometric consistency, the framework does not yet model the view-dependent variations of other LiDAR properties such as ray drop patterns that also influence realism under large viewpoint shifts.
Future work will extend the framework to additional object categories and enhance the physical realism of the synthesized data.

\bibliographystyle{IEEEtran}
\bibliography{References}

@article{geiger2013vision,
  title={Vision meets robotics: The kitti dataset},
  author={Geiger, Andreas and Lenz, Philip and Stiller, Christoph and Urtasun, Raquel},
  journal={The international journal of robotics research},
  volume={32},
  number={11},
  pages={1231--1237},
  year={2013},
  publisher={Sage Publications Sage UK: London, England}
}

@inproceedings{sun2020scalability,
  title={Scalability in perception for autonomous driving: Waymo open dataset},
  author={Sun, Pei and Kretzschmar, Henrik and Dotiwalla, Xerxes and Chouard, Aurelien and Patnaik, Vijaysai and Tsui, Paul and Guo, James and Zhou, Yin and Chai, Yuning and Caine, Benjamin and others},
  booktitle={Proceedings of the IEEE/CVF conference on computer vision and pattern recognition},
  pages={2446--2454},
  year={2020}
}

@inproceedings{caesar2020nuscenes,
  title={nuscenes: A multimodal dataset for autonomous driving},
  author={Caesar, Holger and Bankiti, Varun and Lang, Alex H and Vora, Sourabh and Liong, Venice Erin and Xu, Qiang and Krishnan, Anush and Pan, Yu and Baldan, Giancarlo and Beijbom, Oscar},
  booktitle={Proceedings of the IEEE/CVF conference on computer vision and pattern recognition},
  pages={11621--11631},
  year={2020}
}

@inproceedings{manivasagam2020lidarsim,
  title={Lidarsim: Realistic lidar simulation by leveraging the real world},
  author={Manivasagam, Sivabalan and Wang, Shenlong and Wong, Kelvin and Zeng, Wenyuan and Sazanovich, Mikita and Tan, Shuhan and Yang, Bin and Ma, Wei-Chiu and Urtasun, Raquel},
  booktitle={Proceedings of the IEEE/CVF Conference on Computer Vision and Pattern Recognition},
  pages={11167--11176},
  year={2020}
}

@article{li2022pcgen,
  title={Pcgen: Point cloud generator for lidar simulation},
  author={Li, Chenqi and Ren, Yuan and Liu, Bingbing},
  journal={arXiv preprint arXiv:2210.08738},
  year={2022}
}

@inproceedings{tao2024lidar,
  title={Lidar-nerf: Novel lidar view synthesis via neural radiance fields},
  author={Tao, Tang and Gao, Longfei and Wang, Guangrun and Lao, Yixing and Chen, Peng and Zhao, Hengshuang and Hao, Dayang and Liang, Xiaodan and Salzmann, Mathieu and Yu, Kaicheng},
  booktitle={Proceedings of the 32nd ACM International Conference on Multimedia},
  pages={390--398},
  year={2024}
}

@inproceedings{huang2023neural,
  title={Neural lidar fields for novel view synthesis},
  author={Huang, Shengyu and Gojcic, Zan and Wang, Zian and Williams, Francis and Kasten, Yoni and Fidler, Sanja and Schindler, Konrad and Litany, Or},
  booktitle={Proceedings of the IEEE/CVF International Conference on Computer Vision},
  pages={18236--18246},
  year={2023}
}

@inproceedings{zheng2024lidar4d,
  title={Lidar4d: Dynamic neural fields for novel space-time view lidar synthesis},
  author={Zheng, Zehan and Lu, Fan and Xue, Weiyi and Chen, Guang and Jiang, Changjun},
  booktitle={Proceedings of the IEEE/CVF Conference on Computer Vision and Pattern Recognition},
  pages={5145--5154},
  year={2024}
}

@inproceedings{wu2024dynamic,
  title={Dynamic lidar re-simulation using compositional neural fields},
  author={Wu, Hanfeng and Zuo, Xingxing and Leutenegger, Stefan and Litany, Or and Schindler, Konrad and Huang, Shengyu},
  booktitle={Proceedings of the IEEE/CVF Conference on Computer Vision and Pattern Recognition},
  pages={19988--19998},
  year={2024}
}

@inproceedings{yu2023v2x,
  title={V2x-seq: A large-scale sequential dataset for vehicle-infrastructure cooperative perception and forecasting},
  author={Yu, Haibao and Yang, Wenxian and Ruan, Hongzhi and Yang, Zhenwei and Tang, Yingjuan and Gao, Xu and Hao, Xin and Shi, Yifeng and Pan, Yifeng and Sun, Ning and others},
  booktitle={Proceedings of the IEEE/CVF Conference on Computer Vision and Pattern Recognition},
  pages={5486--5495},
  year={2023}
}

@inproceedings{xu2022v2x,
  title={V2x-vit: Vehicle-to-everything cooperative perception with vision transformer},
  author={Xu, Runsheng and Xiang, Hao and Tu, Zhengzhong and Xia, Xin and Yang, Ming-Hsuan and Ma, Jiaqi},
  booktitle={European conference on computer vision},
  pages={107--124},
  year={2022},
  organization={Springer}
}

@article{li2022v2x,
  title={V2X-Sim: Multi-agent collaborative perception dataset and benchmark for autonomous driving},
  author={Li, Yiming and Ma, Dekun and An, Ziyan and Wang, Zixun and Zhong, Yiqi and Chen, Siheng and Feng, Chen},
  journal={IEEE Robotics and Automation Letters},
  volume={7},
  number={4},
  pages={10914--10921},
  year={2022},
  publisher={IEEE}
}

@inproceedings{dosovitskiy2017carla,
  title={CARLA: An open urban driving simulator},
  author={Dosovitskiy, Alexey and Ros, German and Codevilla, Felipe and Lopez, Antonio and Koltun, Vladlen},
  booktitle={Conference on robot learning},
  pages={1--16},
  year={2017},
  organization={PMLR}
}

@inproceedings{yu2022dair,
  title={Dair-v2x: A large-scale dataset for vehicle-infrastructure cooperative 3d object detection},
  author={Yu, Haibao and Luo, Yizhen and Shu, Mao and Huo, Yiyi and Yang, Zebang and Shi, Yifeng and Guo, Zhenglong and Li, Hanyu and Hu, Xing and Yuan, Jirui and others},
  booktitle={Proceedings of the IEEE/CVF conference on computer vision and pattern recognition},
  pages={21361--21370},
  year={2022}
}

@inproceedings{hao2024rcooper,
  title={Rcooper: A real-world large-scale dataset for roadside cooperative perception},
  author={Hao, Ruiyang and Fan, Siqi and Dai, Yingru and Zhang, Zhenlin and Li, Chenxi and Wang, Yuntian and Yu, Haibao and Yang, Wenxian and Yuan, Jirui and Nie, Zaiqing},
  booktitle={Proceedings of the IEEE/CVF conference on computer vision and pattern recognition},
  pages={22347--22357},
  year={2024}
}

@inproceedings{yuan2018pcn,
  title={Pcn: Point completion network},
  author={Yuan, Wentao and Khot, Tejas and Held, David and Mertz, Christoph and Hebert, Martial},
  booktitle={2018 international conference on 3D vision (3DV)},
  pages={728--737},
  year={2018},
  organization={IEEE}
}

@article{vaswani2017attention,
  title={Attention is all you need},
  author={Vaswani, Ashish and Shazeer, Noam and Parmar, Niki and Uszkoreit, Jakob and Jones, Llion and Gomez, Aidan N and Kaiser, {\L}ukasz and Polosukhin, Illia},
  journal={Advances in neural information processing systems},
  volume={30},
  year={2017}
}

@inproceedings{huang2020pf,
  title={Pf-net: Point fractal network for 3d point cloud completion},
  author={Huang, Zitian and Yu, Yikuan and Xu, Jiawen and Ni, Feng and Le, Xinyi},
  booktitle={Proceedings of the IEEE/CVF conference on computer vision and pattern recognition},
  pages={7662--7670},
  year={2020}
}

@inproceedings{yu2021pointr,
  title={Pointr: Diverse point cloud completion with geometry-aware transformers},
  author={Yu, Xumin and Rao, Yongming and Wang, Ziyi and Liu, Zuyan and Lu, Jiwen and Zhou, Jie},
  booktitle={Proceedings of the IEEE/CVF international conference on computer vision},
  pages={12498--12507},
  year={2021}
}

@inproceedings{yan2025symmcompletion,
  title={SymmCompletion: High-Fidelity and High-Consistency Point Cloud Completion with Symmetry Guidance},
  author={Yan, Hongyu and Li, Zijun and Luo, Kunming and Lu, Li and Tan, Ping},
  booktitle={Proceedings of the AAAI Conference on Artificial Intelligence},
  volume={39},
  number={9},
  pages={9094--9102},
  year={2025}
}

@inproceedings{shi2020pv,
  title={Pv-rcnn: Point-voxel feature set abstraction for 3d object detection},
  author={Shi, Shaoshuai and Guo, Chaoxu and Jiang, Li and Wang, Zhe and Shi, Jianping and Wang, Xiaogang and Li, Hongsheng},
  booktitle={Proceedings of the IEEE/CVF conference on computer vision and pattern recognition},
  pages={10529--10538},
  year={2020}
}

@article{liu2023large,
  title={Large-scale LiDAR consistent mapping using hierarchical LiDAR bundle adjustment},
  author={Liu, Xiyuan and Liu, Zheng and Kong, Fanze and Zhang, Fu},
  journal={IEEE Robotics and Automation Letters},
  volume={8},
  number={3},
  pages={1523--1530},
  year={2023},
  publisher={IEEE}
}

@inproceedings{lang2019pointpillars,
  title={Pointpillars: Fast encoders for object detection from point clouds},
  author={Lang, Alex H and Vora, Sourabh and Caesar, Holger and Zhou, Lubing and Yang, Jiong and Beijbom, Oscar},
  booktitle={Proceedings of the IEEE/CVF conference on computer vision and pattern recognition},
  pages={12697--12705},
  year={2019}
}

@inproceedings{mao2022dolphins,
  title={Dolphins: Dataset for collaborative perception enabled harmonious and interconnected self-driving},
  author={Mao, Ruiqing and Guo, Jingyu and Jia, Yukuan and Sun, Yuxuan and Zhou, Sheng and Niu, Zhisheng},
  booktitle={Proceedings of the Asian Conference on Computer Vision},
  pages={4361--4377},
  year={2022}
}

@inproceedings{wang2024deepaccident,
  title={Deepaccident: A motion and accident prediction benchmark for v2x autonomous driving},
  author={Wang, Tianqi and Kim, Sukmin and Wenxuan, Ji and Xie, Enze and Ge, Chongjian and Chen, Junsong and Li, Zhenguo and Luo, Ping},
  booktitle={Proceedings of the AAAI Conference on Artificial Intelligence},
  volume={38},
  number={6},
  pages={5599--5606},
  year={2024}
}

@inproceedings{li2023proxyformer,
  title={Proxyformer: Proxy alignment assisted point cloud completion with missing part sensitive transformer},
  author={Li, Shanshan and Gao, Pan and Tan, Xiaoyang and Wei, Mingqiang},
  booktitle={Proceedings of the IEEE/CVF conference on computer vision and pattern recognition},
  pages={9466--9475},
  year={2023}
}

@inproceedings{fu2023vapcnet,
  title={VAPCNet: viewpoint-aware 3D point cloud completion},
  author={Fu, Zhiheng and Wang, Longguang and Xu, Lian and Wang, Zhiyong and Laga, Hamid and Guo, Yulan and Boussaid, Farid and Bennamoun, Mohammed},
  booktitle={Proceedings of the IEEE/CVF International Conference on Computer Vision},
  pages={12108--12118},
  year={2023}
}

@inproceedings{zhou2022seedformer,
  title={Seedformer: Patch seeds based point cloud completion with upsample transformer},
  author={Zhou, Haoran and Cao, Yun and Chu, Wenqing and Zhu, Junwei and Lu, Tong and Tai, Ying and Wang, Chengjie},
  booktitle={European conference on computer vision},
  pages={416--432},
  year={2022},
  organization={Springer}
}

@inproceedings{xie2020grnet,
  title={Grnet: Gridding residual network for dense point cloud completion},
  author={Xie, Haozhe and Yao, Hongxun and Zhou, Shangchen and Mao, Jiageng and Zhang, Shengping and Sun, Wenxiu},
  booktitle={European conference on computer vision},
  pages={365--381},
  year={2020},
  organization={Springer}
}

@inproceedings{wang2021voxel,
  title={Voxel-based network for shape completion by leveraging edge generation},
  author={Wang, Xiaogang and Ang, Marcelo H and Lee, Gim Hee},
  booktitle={Proceedings of the IEEE/CVF international conference on computer vision},
  pages={13189--13198},
  year={2021}
}

@inproceedings{huang2021rfnet,
  title={RFNet: Recurrent forward network for dense point cloud completion},
  author={Huang, Tianxin and Zou, Hao and Cui, Jinhao and Yang, Xuemeng and Wang, Mengmeng and Zhao, Xiangrui and Zhang, Jiangning and Yuan, Yi and Xu, Yifan and Liu, Yong},
  booktitle={Proceedings of the IEEE/CVF international conference on computer vision},
  pages={12508--12517},
  year={2021}
}

@inproceedings{liu2020morphing,
  title={Morphing and sampling network for dense point cloud completion},
  author={Liu, Minghua and Sheng, Lu and Yang, Sheng and Shao, Jing and Hu, Shi-Min},
  booktitle={Proceedings of the AAAI conference on artificial intelligence},
  volume={34},
  number={07},
  pages={11596--11603},
  year={2020}
}

@inproceedings{pan2021variational,
  title={Variational relational point completion network},
  author={Pan, Liang and Chen, Xinyi and Cai, Zhongang and Zhang, Junzhe and Zhao, Haiyu and Yi, Shuai and Liu, Ziwei},
  booktitle={Proceedings of the IEEE/CVF conference on computer vision and pattern recognition},
  pages={8524--8533},
  year={2021}
}

@inproceedings{wang2020cascaded,
  title={Cascaded refinement network for point cloud completion},
  author={Wang, Xiaogang and Ang Jr, Marcelo H and Lee, Gim Hee},
  booktitle={Proceedings of the IEEE/CVF conference on computer vision and pattern recognition},
  pages={790--799},
  year={2020}
}

@inproceedings{tchapmi2019topnet,
  title={Topnet: Structural point cloud decoder},
  author={Tchapmi, Lyne P and Kosaraju, Vineet and Rezatofighi, Hamid and Reid, Ian and Savarese, Silvio},
  booktitle={Proceedings of the IEEE/CVF conference on computer vision and pattern recognition},
  pages={383--392},
  year={2019}
}



\begin{IEEEbiography}[
{\includegraphics[width=1in,height=1.25in,clip,keepaspectratio]{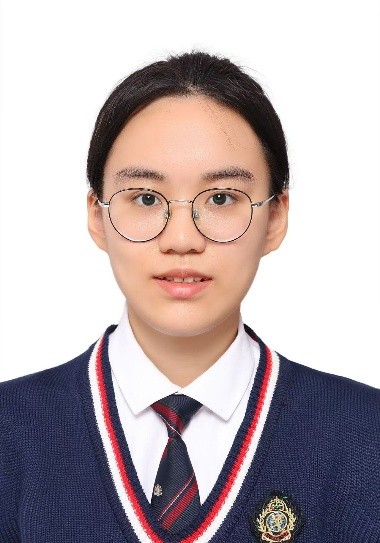}}]
  {Yuhan Xia} is pursuing the B.S. degree at the School of Automation and Intelligent Sensing, Shanghai Jiao Tong University, Shanghai, China. Her research interest is in 3D scene reconstruction for autonomous driving.
\end{IEEEbiography}

\begin{IEEEbiography}[
{\includegraphics[width=1in,height=1.25in,clip,keepaspectratio]{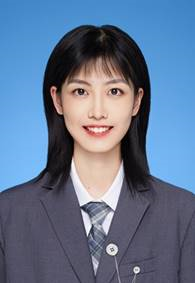}}]
  {Runxin Zhao} received the B.Eng. degree from Xi'an Jiao Tong University, Shannxi, China, in 2023. She is currently a Ph.D. student in control science and engineering at the Department of Automation, Shanghai Jiao Tong University, Shanghai, China. Her research interest is in autonomous vehicle localization and LiDAR perception.
\end{IEEEbiography}

\begin{IEEEbiography}[
{\includegraphics[width=1in,height=1.25in,clip,keepaspectratio]{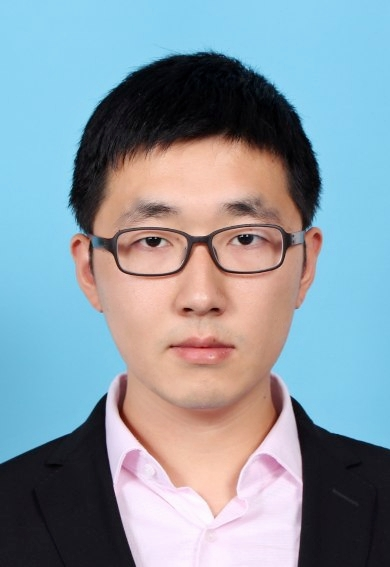}}]
  {Hanyang Zhuang} received the Ph.D. degree from Shanghai Jiao Tong University, Shanghai, China, in 2018. He has worked as a postdoctoral researcher at Shanghai Jiao Tong University from 2020 to 2022. He is currently an assistant research professor at Shanghai Jiao Tong University implementing research works related to intelligent vehicles. His research interest is in autonomous driving and cooperative driving systems.
\end{IEEEbiography}

\begin{IEEEbiography}[
  {\includegraphics[width=1in,height=1.25in,clip,keepaspectratio]{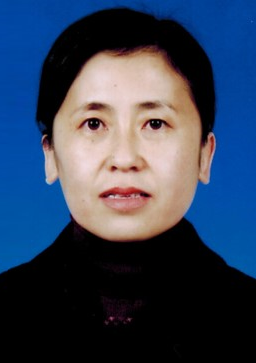}}]
  {Chunxiang Wang} received a Ph.D. degree in mechanical engineering from the Harbin Institute of Technology, China, in 1999. She is currently an Associate Professor at the Department of Automation, Shanghai Jiao Tong University, Shanghai, China. She has been working in the field of intelligent vehicles for more than ten years and participated in several related research projects, such as the European CyberC3 Project and ITER Transfer Cask project. Her research interests include intelligent driving, assistant driving, and mobile robots.
\end{IEEEbiography}

\begin{IEEEbiography}[
  {\includegraphics[width=1in,height=1.25in,clip,keepaspectratio]{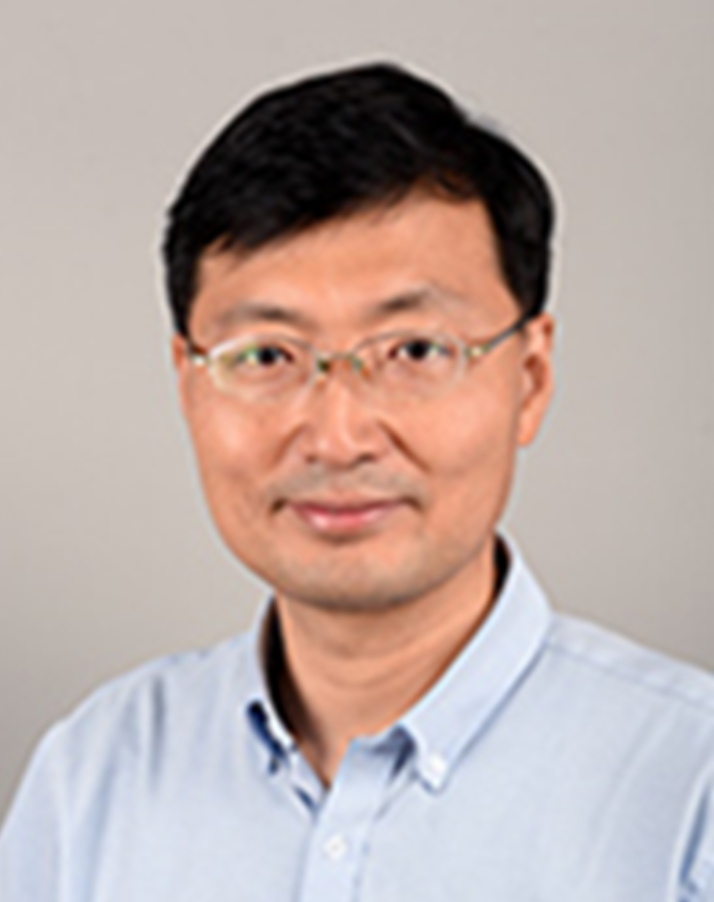}}]
  {Ming Yang} received Master's and Ph.D. degrees from Tsinghua University, Beijing, China, in 1999 and 2003, respectively. He is currently a Full Tenure Professor at Shanghai Jiao Tong University, director of the Department of Automation, and the deputy director of the Innovation Center of Intelligent Connected Vehicles. He has been working in the field of intelligent vehicles for more than 20 years. Furthermore, he participated in several related research projects, such as the THMR-V project (first intelligent vehicle in China), European CyberCars and CyberMove projects, CyberC3 project, ITER transfer cask project, etc.
\end{IEEEbiography}

\end{document}